\documentclass[11pt,letterpaper]{article}

\usepackage[T1]{fontenc}
\usepackage{newtxtext}
\usepackage{newtxmath}
\usepackage[margin=1in]{geometry}
\usepackage[hyphens]{url}
\usepackage{natbib}
\usepackage{booktabs}
\usepackage{array}
\usepackage{colortbl}
\usepackage{graphicx}
\usepackage{amsmath}
\usepackage{microtype}
\usepackage{placeins}
\usepackage{xstring}
\usepackage[hidelinks]{hyperref}

\graphicspath{{figures/}}

\let\ArxivOriginalIncludeGraphics\includegraphics
\renewcommand{\includegraphics}[2][]{%
  \IfStrEq{#2}{v543_real_city_analogues.pdf}{%
    \ArxivOriginalIncludeGraphics[width=.58\textwidth,keepaspectratio]{#2}%
  }{%
    \IfStrEq{#2}{candidates/v530_fig3_two_panel_candidate.pdf}{%
      \ArxivOriginalIncludeGraphics[width=.56\textwidth,keepaspectratio]{#2}%
    }{%
      \IfStrEq{#2}{v528_joint_profile_validation.pdf}{%
        \ArxivOriginalIncludeGraphics[width=.55\textwidth,keepaspectratio]{#2}%
      }{%
        \IfStrEq{#2}{v527_40_dimension_regional_domains_appendix.pdf}{%
          \ArxivOriginalIncludeGraphics[width=.72\textwidth,keepaspectratio]{#2}%
        }{%
          \ArxivOriginalIncludeGraphics[#1]{#2}%
        }%
      }%
    }%
  }%
}

\newcommand{\ArxivTitle}{%
  Mapping the City Through the Lens of Language Models%
}

\title{\ArxivTitle}
\author{%
  Wanqi Liu\textsuperscript{1,\ensuremath{\dagger}} \quad
  Rong Zhao\textsuperscript{1,\ensuremath{\dagger},*} \quad
  Zhizhou Sha\textsuperscript{3} \quad
  Qinyu Cui\textsuperscript{4} \quad
  Yecheng Zhang\textsuperscript{2}
  \\[0.8em]
  \small\textsuperscript{1}Centre for Advanced Spatial Analysis (CASA),
  University College London, London, UK
  \\
  \small\textsuperscript{2}School of Architecture, Tsinghua University,
  Beijing, China
  \\
  \small\textsuperscript{3}Department of Computer Science,
  The University of Texas at Austin, Austin, TX, USA
  \\
  \small\textsuperscript{4}Department of Civil and Transportation Engineering,
  South China University of Technology, Guangzhou, China
  \\[0.45em]
  \small\textsuperscript{\ensuremath{\dagger}}Equal contribution.
  \quad
  \textsuperscript{*}Corresponding author: \href{mailto:rong.zhao.25@ucl.ac.uk}{rong.zhao.25@ucl.ac.uk}
}
\date{}

\newcommand{\ArxivAbstract}{%
Language models often complete an underspecified reference to ``a city'' with
unstated assumptions about urban size, form, infrastructure, environment, and
function. We measure those assumptions without naming places. Ten open-weight
checkpoints rate anonymized profiles derived from real morphological urban
centres across 40 audited indicators and seven domains. The design combines
constrained probability-based ratings, prespecified reliability screens,
lineage-aware aggregation, multiple population weightings, an independent
replication sample, and whole-profile validation. The clearest shared tendency
favours urban profiles with larger developed area, faster recent growth,
greater mapped infrastructure and non-residential capacity, and less sparse
form. Most eligible directions recur in the replication data, and direct
ratings of complete profiles show moderate agreement with the indicator-wise
construction. Geographic differences shrink after accounting for city scale
and development, while reliably measured paired tasks indicate that typicality
and desirability are often closely aligned. The framework makes an otherwise
vague notion of what models regard as an ordinary city empirically traceable.
The resulting evidence delineates a shared yet model-dependent portrait of the
city through the lens of language models.%
}

\newcommand{\ArxivSourceLedger}{}
\newcommand{\ArxivMeasurementLedger}{}



\hypersetup{
  pdftitle={Mapping the City Through the Lens of Language Models},
  pdfauthor={Wanqi Liu, Rong Zhao, Zhizhou Sha, Qinyu Cui, Yecheng Zhang}
}

\begin{document}
\bibliographystyle{plainnat}
\maketitle
\begin{quote}
  \small
  \noindent\textbf{Abstract.} \ArxivAbstract
\end{quote}
\section{Introduction}

Language models routinely resolve underspecified queries by supplying unstated
context from their learned distributions. In geographic settings, this
behavior is consequential because the category \emph{city} spans substantial
variation in scale, form, infrastructure, ecology, function, and development.
When a prompt refers only to ``a city,'' a model must still privilege some
combinations of these properties. We ask which anonymous urban profiles models
treat as central examples of the category.

Existing geographic evaluations predominantly examine factual knowledge, task
performance, or representational disparities associated with named locations
\citep{faisal2023geographic,manvi2024geographic,dunn2024populations,
li2024land}. Urban benchmarks similarly evaluate planning, prediction, and
reasoning in a predefined set of cities
\citep{feng2024citybench,zheng2025urbanplanbench}. These studies establish that
model behavior varies geographically without identifying the urban profile
elicited as category-central when place identity is absent.
\citet{campanella2024bigcity} document metropolitan-size bias in a specific
labor-market task; we generalize that question from task performance to
multidimensional judgments of \emph{city}.

A direct generation prompt is poorly suited to this problem. Asking a model to
describe a ``typical city'' conflates category centrality with linguistic
style, memorized places, cultural familiarity, and normative ideas about
desirable development and leaves the denominator undefined. We therefore
define the \emph{Default City} as a behavioral ranking over empirical urban
profiles, conditional on a specified model, elicitation task, and reference
population. City-count, resident-weighted, and region-balanced typicality are
distinct estimands.

\begin{figure*}[!t]
\centering
\includegraphics[width=.99\textwidth]{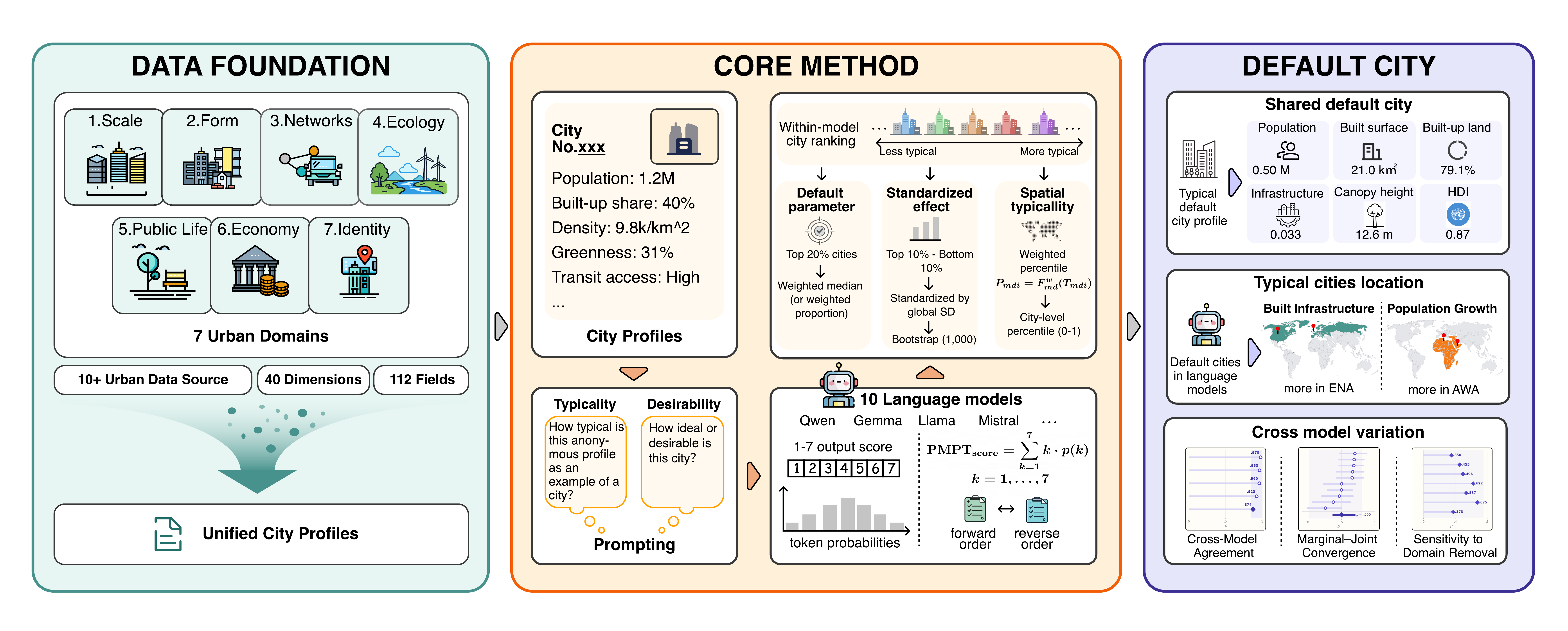}
\caption{Overview of the Default City audit. Heterogeneous urban data are
harmonized into anonymous profiles spanning 40 dimensions and seven conceptual
domains. Ten language-model checkpoints score profile typicality; the
resulting rankings yield default parameters, standardized effects,
geographic affinity, and cross-model comparisons.}
\label{fig:framework}
\end{figure*}

We assemble anonymous profiles of globally harmonized morphological urban
centres across 40 source-audited dimensions and seven conceptual domains. Ten
open-weight checkpoints assign constrained 1--7 typicality ratings using
next-token probabilities. Geographic identifiers and evaluative labels are
excluded. Field-order agreement, response dispersion, reduced-form agreement,
and row-specific reliability diagnostics identify unstable outputs; shared
model lineages are aggregated through dependency components.
Figure~\ref{fig:framework} summarizes the audit.

Raw-unit parameters, standardized contrasts, and city-level percentiles expose
the prototype's content and geography. A disjoint 512-city sample evaluates
replication; direct ratings of complete 40-indicator profiles test whether
marginal judgments cohere at the city level; and matched desirability judgments
probe normative coupling.

The evidence reveals a recurring scale--growth--infrastructure core alongside
checkpoint, geographic, developmental, and normative variation. Our
contributions are a population-conditional formulation of geographic category
centrality; an auditable multidimensional framework combining probabilistic
judgments, provenance, reliability diagnostics, dependency-aware aggregation,
and replication; and a global behavioral characterization of the urban
profiles that language models treat as typical.

\section{Related Work}

\subsection{Geographic Bias and Urban Evaluation}

Geographic audits document uneven knowledge, population representation, and
geopolitical behavior for named locations
\citep{faisal2023geographic,manvi2024geographic,dunn2024populations,
li2024land}. CityBench and UrbanPlanBench evaluate capabilities in predefined
cities \citep{feng2024citybench,zheng2025urbanplanbench}; task-specific and
street-view studies further reveal metropolitan-size and cultural tilts
\citep{campanella2024bigcity,zhao2026urbanperception}. We instead study the
category-central ranking elicited from anonymous, globally sampled real-city
profiles. Complementary work tests whether generated symbolic and perceptual
urban data reproduce urban-science patterns, finding theoretical alignment
alongside limited diversity and systematic parameter deviations
\citep{zhang2025ai4us}.

\subsection{Prototype Structure and Underspecification}

Prototype theories treat category membership as graded
\citep{rosch1975categories,medin1978context,hampton2012thinking}. A prototype
is a central configuration, not necessarily an arithmetic mean. We use this
distinction behaviorally, without assuming human-like conceptual organization.
Underspecification changes model responses and their evaluation
\citep{malaviya2025contextualized}, while population-aware audits show that
alignment depends on the reference distribution \citep{santurkar2023opinions}.
Our three population targets make that distribution explicit.

\subsection{Behavioral Measurement and Audit Reliability}

Behavioral tests expose invariances and failure modes hidden by aggregate
scores \citep{ribeiro2020checklist,liang2023helm}; judgment tasks can also be
sensitive to answer or item order
\citep{wang2024fair,pezeshkpour2024order}. We therefore use constrained
next-token distributions and record order agreement, dispersion, and
reduced-form agreement. Provenance, model pins, eligibility rules, and complete
matrices follow reporting and reproducibility guidance
\citep{gebru2021datasheets,mitchell2019modelcards,pineau2021reproducibility,
pattnayak2026reproevalcard}. Unreliable cells remain descriptive; replication
and direct joint-profile judgments test stability and behavioral coherence.

\section{Method}

\subsection{Data Foundation}

We use morphological urban centres from the official GHS Urban Centre
Database R2024A V1.2, defined by globally harmonized density, population, and
contiguity rules \citep{melchiorri2024stats,dijkstra2021degree}. Of its 11,422
quality-controlled centres, 11,377 form the complete multiview frame. A
region- and population-stratified probability sample supplies 2,000 profiles
and design weights; a disjoint 512-profile sample supports replication.

The source audit harmonizes 40 dimensions into seven organizing domains:
\emph{Scale, Extent \& Change}; \emph{Built Form \& Land Use};
\emph{Networks, Mobility \& Infrastructure}; \emph{Ecology \& Environment};
\emph{Population, Housing \& Public Life}; \emph{Economy, Function \&
Centrality}; and \emph{Culture, Institutions \& Identity}. Thirty-six
dimensions share the 2,000-profile frame. Three Global Streetscapes
dimensions use its complete 356-city panel, and one GlobalBuildingAtlas
dimension uses the complete 353-city GBA--Streetscapes joint panel
\citep{bondarenko2025worldpop,overture2026data,hou2024globalstreetscapes,
zhu2025globalbuildingatlas,mobilitydata2026gtfs}. Prompts omit geographic and
source identifiers. The Supplement records missingness, 13 directly
materialized products in eight source families, inherited thematic lineages,
versions, and receipts.

\ArxivSourceLedger

\subsection{Behavioral Elicitation}

For model $m$, dimension $d$, profile $i$, and field order $o$, let
$p_{mdio}(k)$ be the constrained next-token probability of rating
$k\in\{1,\ldots,7\}$. We use the probability-weighted rating
\begin{equation}
r_{mdio}=\sum_{k=1}^{7}k\,p_{mdio}(k).
\end{equation}
Single-field dimensions are scored once. Multi-field dimensions use a
hash-selected cyclic offset and the exact reverse order at the same offset;
their mean. The hash depends only on dimension and anonymous profile ID.
Temperature is zero, generation is at most one token, and fields retain
physical units without evaluative labels. The prompt asks how typical each
record is of what people mean by a city; matched desirability prompts form a
separate diagnostic.

Ten locally served, version-pinned checkpoints rate all 40 dimensions:
Qwen3-8B, Qwen2.5-7B-Instruct, DeepSeek-R1-Distill-Qwen-7B, Gemma-2-9B-IT,
Gemma-3-12B-IT, Llama-3.1-8B-Instruct, Mistral-Nemo-12B,
OLMo-2-7B-Instruct, Phi-4-14B, and GLM-4-9B-0414
\citep{yang2025qwen3,gemmateam2024gemma2,dubey2024llama3,
mistral2024nemo,olmo2025furious}. Grouping Qwen-derived and Gemma checkpoints
by lineage leaves seven dependency components.

\subsection{Measurement and Aggregation}

Fully calibrated cells are clean when valid responses are at least .99,
typicality standard deviation at least .15, forward/reverse rank correlation
at least .80 for multi-field dimensions, and reduced/full-form correlation at
least .85 where applicable. Source-specific rows use matched response,
dispersion, order, coverage, and sample-size diagnostics. The complete
$40\times10$ matrix retains all clean, low-dispersion, and
order/reduction-sensitive cells; standardized feature contrasts use clean
cells.

Within each model--dimension cell that meets its row-specific clean criterion,
we compare every source feature between the highest and lowest weighted
deciles of typicality.
For target $t$,
\begin{equation}
\Delta_{mdft} =
\frac{\bar{x}_{f,t}(T_{md}\ge q_{.9})-
\bar{x}_{f,t}(T_{md}\le q_{.1})}{s_{f,t}}.
\end{equation}
The primary city-count target uses sampling weights. A population-weighted
target additionally weights by residents, and a region-balanced target gives
each world region equal weight; these are separate target populations.

For raw-unit interpretation, a model's Default is the weighted median within
the upper typicality quintile (a weighted share for the binary capital
indicator). Combined values median eligible checkpoints within dependency
components and then median components. Uncertainty uses 1,000
Rao--Wu--Yue--Beaumont survey bootstrap replicates that recompute weights,
thresholds, and contrasts \citep{rao1992resampling,beaumont2012bootstrap};
main and replication seeds are 20260724/20260725. Correlated indicators
preclude causal interpretation.

\subsection{Comparative and Validation Analyses}

For descriptive geography and agreement, each eligible checkpoint--dimension
score becomes a design-weighted percentile; rankings are medianed within
dependency components and, for maps, across components. Coordinates are
joined only after inference. Regional affinity is the region mean minus the
row-specific global median. Figure~\ref{fig:regional-affinity} first aggregates
dimensions within domains, then gives domains and dependency components equal
weight. Its conditioned synthesis uses 123 reliable non-scale cells and
residualizes percentiles on quadratic ranks for population, built surface,
and density, with and without subnational HDI.

Checkpoint heterogeneity is the design-weighted mean absolute profile-rank
difference from a consensus that excludes the checkpoint's entire dependency
component. Table~\ref{tab:model-divergence} summarizes this distance across
the 40 dimensions. Scale-conditioned contrasts re-rank non-scale dimensions
within world-region and population-band strata before aggregation. Matched
typicality and desirability are compared on 256 profiles only when both
constructs meet their response, dispersion, and order criteria. The disjoint
512-profile replication repeats the primary prompts, weights, and analysis
for every eligible direction under all three target populations.

Finally, every centre with complete coverage for one designated parameter in
all 40 dimensions ($n=48$; 29 countries, six regions) receives a direct
anonymous 40-line rating in forward and reverse order, averaged as $J_{mi}$.
The marginal predictor $M_{mi}$ averages within-dimension midranks first
within the seven domains and then equally across domains. We report
checkpoint-level Spearman $\rho(M_{mi},J_{mi})$, its dependency-balanced
median, component signs, order agreement, and a 5,000-replicate city-bootstrap
interval.

\section{Results}

\subsection{The Multidimensional Default-City Portrait}

Table~\ref{tab:model-results} makes the central empirical asymmetry visible:
where typicality is measurable, the Default City lies well above the
row-specific global centre on scale, built extent, mapped infrastructure,
human development, everyday services, and non-residential capacity. The
dependency-balanced portrait has 0.50 million residents, 21.0 km$^2$ of built
surface, 79.1\% built land, an infrastructure index of .033, 12.6 m canopy
height, and subnational HDI of .87. Most of these values lie in the
81st--90th percentiles of their own world or availability frames. These
row-wise summaries describe the portrait dimension by dimension.

\begin{table*}[!htbp]
\centering
\begingroup
\footnotesize
\setlength{\tabcolsep}{0.75pt}
\renewcommand{\arraystretch}{1.02}
\begin{tabular*}{\textwidth}{@{\extracolsep{\fill}}>{\raggedright\arraybackslash}p{.215\textwidth}r*{10}{r}rr@{}}
\toprule
\textbf{Parameter} & \textbf{Global} & \multicolumn{10}{c}{\textbf{Model-specific Default}} & \textbf{Combined} & \textbf{Percentile} \\
\cmidrule(lr){3-12}
 &  & \multicolumn{1}{c}{\shortstack{Qwen\\3}} & \multicolumn{1}{c}{\shortstack{Qwen\\2.5}} & \multicolumn{1}{c}{\shortstack{Deep\\Seek}} & \multicolumn{1}{c}{\shortstack{Gemma\\2}} & \multicolumn{1}{c}{\shortstack{Gemma\\3}} & \multicolumn{1}{c}{\shortstack{Llama\\3.1}} & \multicolumn{1}{c}{\shortstack{Mistral}} & \multicolumn{1}{c}{\shortstack{OLMo\\2}} & \multicolumn{1}{c}{\shortstack{Phi\\4}} & \multicolumn{1}{c}{\shortstack{GLM\\4}} &  &  \\
\midrule
\multicolumn{14}{@{}l}{\textbf{Scale, Extent \& Change}} \\
\rule{0pt}{2.35ex}Population (millions) & 0.11 & 0.50 & 0.50 & 0.07 & 0.50 & 0.28 & 0.49 & \cellcolor[gray]{0.92}0.50 & \cellcolor[gray]{0.92}0.47 & 0.50 & 0.50 & \textbf{0.50} & 90 \\
\rule{0pt}{2.35ex}Built surface (km$^2$) & 4.5 & 21.2 & 21.6 & \cellcolor[gray]{0.92}7.6 & 20.8 & 21.2 & \cellcolor[gray]{0.92}18.8 & 19.5 & \cellcolor[gray]{0.92}16.5 & 21.0 & \cellcolor[gray]{0.92}21.0 & \textbf{21.0} & 90 \\
\rule{0pt}{2.35ex}Footprint compactness (0--1) & 0.42 & 0.54 & 0.54 & \cellcolor[gray]{0.92}0.49 & 0.54 & 0.55 & \cellcolor[gray]{0.92}0.54 & \cellcolor[gray]{0.92}0.54 & 0.54 & 0.54 & \cellcolor[gray]{0.92}0.54 & \textbf{0.54} & 86 \\
\rule{0pt}{2.35ex}Population growth (\%/yr) & 0.3 & 1.7 & 2.2 & \cellcolor[gray]{0.92}0.5 & 2.2 & 2.2 & 2.2 & 2.2 & 2.2 & 2.2 & \cellcolor[gray]{0.92}-0.7 & \textbf{2.2} & 90 \\
\rule{0pt}{2.35ex}Pop.--built growth gap (pp/yr) & -0.4 & 1.1 & 1.4 & \cellcolor[gray]{0.92}0.4 & 1.1 & 1.3 & 1.2 & 1.3 & 0.8 & 1.4 & \cellcolor[gray]{0.92}1.4 & \textbf{1.2} & 87 \\
\midrule
\multicolumn{14}{@{}l}{\textbf{Built Form \& Land Use}} \\
\rule{0pt}{2.35ex}Compact LCZ share (\%) & 5.6 & \cellcolor[gray]{0.92}14.8 & \cellcolor[gray]{0.92}7.9 & \cellcolor[gray]{0.92}7.9 & 17.5 & \cellcolor[gray]{0.92}18.7 & 13.5 & 9.2 & \cellcolor[gray]{0.92}12.7 & 2.9 & 4.6 & \textbf{9.2} & 62 \\
\rule{0pt}{2.35ex}Built land share (\%) & 28.3 & 76.3 & 78.4 & \cellcolor[gray]{0.92}60.0 & 68.8 & 83.5 & 83.2 & \cellcolor[gray]{0.92}83.0 & \cellcolor[gray]{0.92}59.9 & 79.1 & 83.9 & \textbf{79.1} & 86 \\
\rule{0pt}{2.35ex}Height/street-width ratio & 0.29 & 0.46 & 0.43 & \cellcolor[gray]{0.92}0.34 & \cellcolor[gray]{0.92}0.46 & \cellcolor[gray]{0.92}0.18 & \cellcolor[gray]{0.92}0.46 & \cellcolor[gray]{0.92}0.46 & \cellcolor[gray]{0.92}0.36 & 0.46 & \cellcolor[gray]{0.92}0.46 & \textbf{0.45} & 89 \\
\midrule
\multicolumn{14}{@{}l}{\textbf{Networks, Mobility \& Infrastructure}} \\
\rule{0pt}{2.35ex}Infrastructure index & 0.013 & 0.036 & 0.032 & \cellcolor[gray]{0.92}0.032 & 0.031 & 0.033 & 0.031 & 0.034 & 0.033 & 0.033 & 0.033 & \textbf{0.033} & 85 \\
\rule{0pt}{2.35ex}Download speed (Mbps) & 79 & \cellcolor[gray]{0.92}238 & \cellcolor[gray]{0.92}297 & \cellcolor[gray]{0.92}153 & \cellcolor[gray]{0.92}275 & 281 & \cellcolor[gray]{0.92}262 & 275 & \cellcolor[gray]{0.92}272 & 289 & \cellcolor[gray]{0.92}209 & \textbf{281} & 84 \\
\midrule
\multicolumn{14}{@{}l}{\textbf{Ecology \& Environment}} \\
\rule{0pt}{2.35ex}Canopy height (m) & 8.6 & 13.2 & 13.4 & \cellcolor[gray]{0.92}7.8 & 9.6 & 13.9 & 13.3 & 13.9 & 9.7 & 12.6 & 12.4 & \textbf{12.6} & 81 \\
\rule{0pt}{2.35ex}Mean temperature ($^\circ$C) & 21.8 & \cellcolor[gray]{0.92}25.2 & \cellcolor[gray]{0.92}25.7 & \cellcolor[gray]{0.92}22.8 & \cellcolor[gray]{0.92}17.1 & 26.2 & \cellcolor[gray]{0.92}23.4 & \cellcolor[gray]{0.92}25.3 & 26.6 & \cellcolor[gray]{0.92}24.1 & 26.6 & \textbf{26.6} & 84 \\
\rule{0pt}{2.35ex}High-green population (\%) & 2.7 & 26.6 & 26.6 & \cellcolor[gray]{0.92}7.7 & 24.7 & 26.0 & 26.6 & 26.6 & 26.6 & 25.2 & 26.6 & \textbf{26.6} & 90 \\
\rule{0pt}{2.35ex}Visible vegetation (\%) & 14.2 & 25.9 & 25.7 & 7.9 & \cellcolor[gray]{0.92}26.0 & 9.5 & 24.3 & 26.1 & 25.9 & 25.2 & \cellcolor[gray]{0.92}24.6 & \textbf{25.5} & 88 \\
\midrule
\multicolumn{14}{@{}l}{\textbf{Population, Housing \& Public Life}} \\
\rule{0pt}{2.35ex}Human development (0--1) & 0.71 & 0.87 & 0.87 & \cellcolor[gray]{0.92}0.76 & 0.87 & 0.87 & 0.87 & 0.87 & 0.87 & 0.87 & \cellcolor[gray]{0.92}0.87 & \textbf{0.87} & 90 \\
\rule{0pt}{2.35ex}Everyday-service share (\%) & 65.4 & 80.0 & 80.0 & \cellcolor[gray]{0.92}66.8 & 80.0 & 80.0 & 80.0 & 80.0 & 80.0 & 80.0 & 80.0 & \textbf{80.0} & 90 \\
\midrule
\multicolumn{14}{@{}l}{\textbf{Economy, Function \& Centrality}} \\
\rule{0pt}{2.35ex}Non-res. volume share (\%) & 3.0 & 23.8 & 22.6 & \cellcolor[gray]{0.92}18.5 & 23.5 & 23.7 & 23.4 & 23.5 & 22.9 & 22.9 & 23.3 & \textbf{23.3} & 89 \\
\rule{0pt}{2.35ex}Functional/centre area ratio & 2.89 & \cellcolor[gray]{0.92}5.75 & 5.24 & \cellcolor[gray]{0.92}4.42 & 6.52 & 5.87 & \cellcolor[gray]{0.92}5.45 & 6.36 & 5.92 & 6.58 & 5.90 & \textbf{6.06} & 86 \\
\midrule
\multicolumn{14}{@{}l}{\textbf{Culture, Institutions \& Identity}} \\
\rule{0pt}{2.35ex}National-capital share (\%) & 1.0 & 3.1 & 3.1 & \cellcolor[gray]{0.92}0.0 & \cellcolor[gray]{0.92}3.2 & \cellcolor[gray]{0.92}3.0 & 4.2 & 5.0 & 2.6 & 1.8 & \cellcolor[gray]{0.92}4.2 & \textbf{3.1} & -- \\
\rule{0pt}{2.35ex}Civic/cultural place share (\%) & 1.3 & 3.7 & 3.7 & \cellcolor[gray]{0.92}0.0 & 3.7 & 3.7 & 3.7 & 3.7 & 2.9 & 3.7 & 3.7 & \textbf{3.7} & 90 \\
\bottomrule
\end{tabular*}
\vspace{2pt}
\noindent\parbox{.99\textwidth}{\footnotesize Continuous values are weighted medians in the upper typicality quintile; the capital row is a weighted share. All 200 model values are printed in black. A light-gray background marks a cell below its measurement-reliability criterion; these values remain descriptive. Cell-level status is reported in the appendix. Combined values median clean checkpoints within dependency components and then across components. Percentiles use the row-specific world or availability frame; the binary capital share uses a dash in the percentile column.}
\endgroup

\caption{Twenty-parameter raw-unit results from the 40-dimensional,
ten-model experiment. Each row reports its global benchmark, all ten
model-specific Defaults, a dependency-balanced combined Default, and its
row-specific global percentile. Continuous values are weighted medians among
the upper typicality quintile; national-capital values are weighted shares.
All 200 model values are printed in black; a light-gray background marks a
cell below its measurement-reliability criterion. Shaded cells remain
descriptive and are excluded from the combined value. The appendix reports the
complete 40-dimensional matrix.}
\label{tab:model-results}
\end{table*}

\ArxivMeasurementLedger

\begin{figure*}[!htbp]
\centering
\includegraphics[width=.99\textwidth]{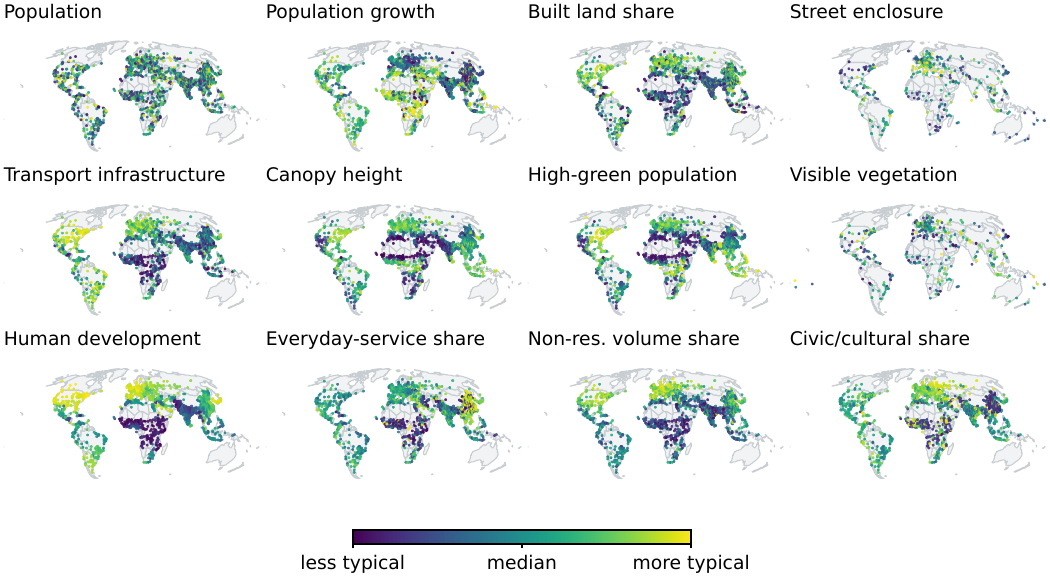}
\caption{City-point view of 12 parameters from
Table~\ref{tab:model-results}. Each point is one sampled or available urban
centre, colored by its typicality percentile on a shared 0--1 scale.
Percentiles are design-weighted and medianed within and across dependency
components. Gray land provides geographic context only; no spatial
interpolation or country-level aggregation is applied.}
\label{fig:global-montage}
\end{figure*}

\begin{figure}[!htb]
\centering
\includegraphics[width=\columnwidth]{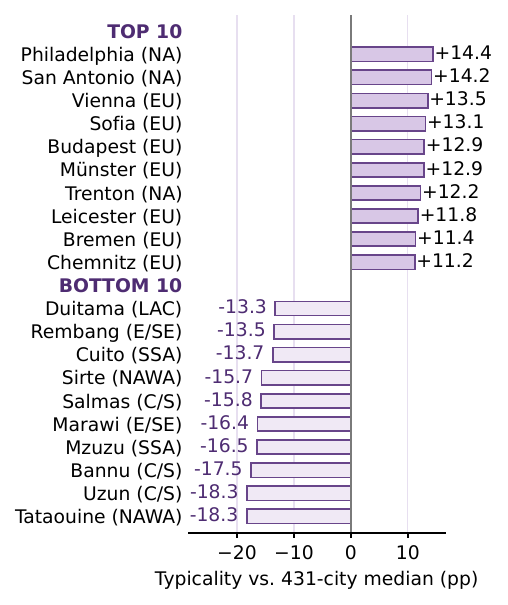}
\caption{Highest- and lowest-ranked real-city analogues. Bars show
seven-domain-balanced typicality relative to the median of 431 centres
observed on all 36 shared-frame dimensions (84 countries). Parentheses denote
the world regions used in Figure~\ref{fig:regional-affinity}.}
\label{fig:city-analogues}
\end{figure}

Domain-balanced aggregation yields a recognizable set of real-city analogues
(Figure~\ref{fig:city-analogues}). Philadelphia and San Antonio lead, followed
by Vienna, Sofia, and Budapest; Tataouine, Uzun, and Bannu occupy the opposite
end. The upper tail is concentrated in Europe and Northern America, whereas
the lower tail spans Northern Africa/Western Asia, Central/Southern Asia,
Sub-Saharan Africa, Eastern/South-Eastern Asia, and Latin America. This
contrast anticipates the regional affinities below.

Among cells meeting their row-specific reliability criterion, high- versus
low-typicality contrasts reinforce the raw-unit portrait. Built extent, built
land, population growth, growth relative to expansion, mapped infrastructure,
canopy, and non-residential capacity are positive across eligible dependency
components. Compact and open LCZ form outrank sparse form, while built growth,
footprint elongation, and several land-cover shares vary across components.
Signs are more stable than magnitudes: built extent and infrastructure occupy
narrow positive ranges, whereas compactness, canopy, and growth relative to
expansion vary more (complete checkpoint table and feature atlas in the
supplement).

Source-specific rows add visual and building context. Visible vegetation and
street-edge enclosure are directionally consistent across checkpoints in the
356-city Streetscapes frame, while height-to-width enclosure is positive in
the 353-city GBA joint frame. The supplement reports all 56 associated field
parameters and their reliability metadata.

\FloatBarrier

\subsection{Geography Reorders the Portrait}

Figure~\ref{fig:global-montage} returns the anonymous-profile rankings to
their withheld locations. Built extent, infrastructure, and non-residential
capacity rank higher across much of Northern America and Europe, whereas
population growth relative to expansion is higher across much of Africa and
Western Asia. The city points expose both regional structure and
within-country variation beyond a single global percentile.

\begin{figure}[!htb]
\centering
\includegraphics[width=\columnwidth]{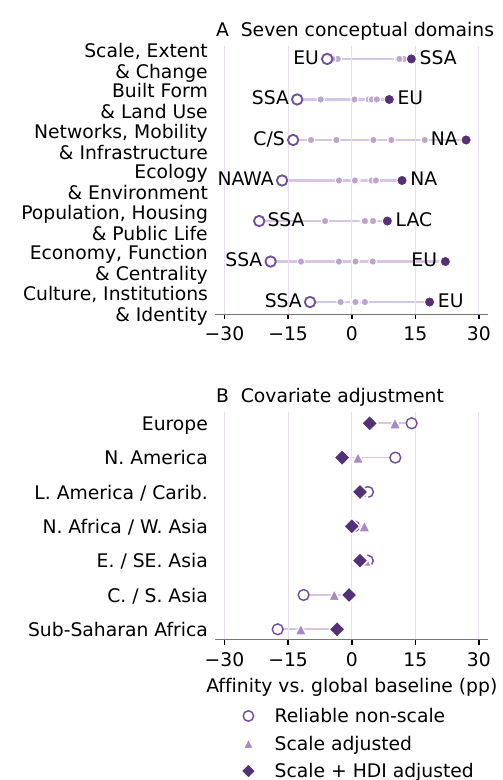}
\caption{Regional affinity relative to each row's global baseline. A:
unadjusted dependency-balanced affinity in each of the seven conceptual
domains. B: synthesis across 123 reliable non-scale cells before adjustment
(circles), after population, built-surface, and density adjustment
(triangles), and after additionally adjusting subnational HDI (diamonds).
The unadjusted all-40 regional synthesis is reported in the text.}
\label{fig:regional-affinity}
\end{figure}

Figure~\ref{fig:regional-affinity} summarizes these patterns by region.
Unadjusted affinity is highest
for Europe ($+10.8$ percentage points) and Northern America ($+10.0$), and
lowest for Sub-Saharan Africa ($-10.2$) and Central/Southern Asia ($-6.5$).
Among reliable non-scale cells, the corresponding unadjusted values are
$+14.1$ and $+10.3$. Population adjustment alone barely changes the regional
ordering (median cell $R^2=.02$). Joint adjustment for population, built
surface, and density reduces Europe to $+10.2$ and Northern America to
$+1.5$ (median $R^2=.17$); adding subnational HDI yields $+4.2$ and $-2.3$
(median $R^2=.28$). Thus population alone does not produce the raw tilt:
scale composition is especially consequential for Northern America, while
development level absorbs much of the remaining Europe--Northern America
contrast. Complete domain and regional results appear in the supplement.

\FloatBarrier

\subsection{Checkpoint Variation around a Shared Core}

Table~\ref{tab:model-divergence} measures checkpoint variation around the
shared portrait. Nine checkpoints form a relatively compact band, with median
divergence between 11.1 and 16.1 percentage points. DeepSeek-R1 is distinct at
23.7, with an IQR of 20.5--28.6.

The pattern agrees with the narrower component-correlation analysis while
adding coverage. Among the 13 dimensions that permit direct
cross-component comparison, the median dimension-level correlation is .874.
Agreement is strongest for non-residential capacity (.978), built extent
(.963), population growth (.960), and infrastructure (.923), but weak for the
two-component building-type view (.156) and moderate for LCZ (.712) and
footprint geometry (.757). Thus the Default City has a shared scale and
infrastructure core alongside substantial variation in several morphological
dimensions.

\subsection{Marginal and Joint Judgments Converge}

\begin{figure}[!htbp]
\centering
\includegraphics[width=\columnwidth]{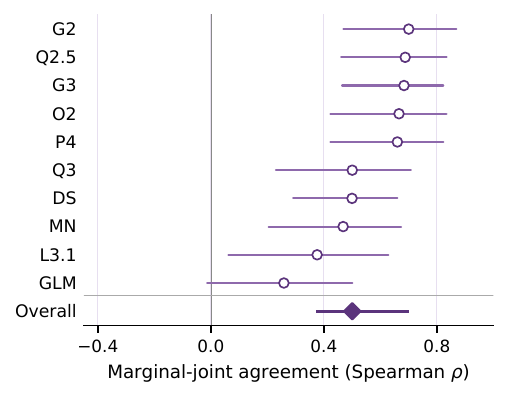}
\caption{Joint-profile convergence across 48 centres. Circles and thin lines
show checkpoint-specific Spearman correlations and 95\% city-bootstrap
intervals between the seven-domain-balanced marginal predictor and direct
40-indicator ratings; the diamond gives the dependency-component-balanced
median.}
\label{fig:joint-profile}
\end{figure}

Figure~\ref{fig:joint-profile} shows positive marginal--joint agreement for
all ten checkpoints ($\rho=.258$--$.700$; checkpoint median .580). After
dependency balancing, the association is
$\rho=.500$ (95\% CI [.378, .695]), with all seven components positive.
Forward/reverse agreement ranges from .502 to .919. Removing any one
conceptual domain leaves the component-balanced association positive
(.350--.675). The result establishes moderate within-model convergence on the
48-city shared-coverage frame.
Sensitivity analyses retain the central rankings across upper-tail thresholds,
aggregation choices, leave-one-component comparisons, and Chinese
instructions ($\rho=.899$--$.990$; Supplement).

\FloatBarrier

\subsection{Replication, Scale, and Desirability}

The replication sample reproduces 285/291 city-count directions, with an
effect-rank correlation of .982. All six reversed directions have intervals
crossing zero, as do 27 additional replication estimates. Across seven
components, city-count direction/rank agreement is .991/.980; population
weighting is similar, while region balancing retains direction and yields
lower rank agreement.

Scale conditioning supplies the strongest qualification to the physical
portrait. After non-scale comparisons are redefined within world region and
population band, all 41 displayed directions remain, but their absolute
magnitudes attenuate by a median 38.1\%. Roads, infrastructure, and
non-residential capacity attenuate most. Together with the conditioned
regional synthesis, this identifies a strong scale component alongside
secondary morphology that persists among more comparable cities.

Typicality and desirability are closely aligned in the 16/60 paired
model--dimension cells with reliable measurement: their rank correlations are
.904--.997 (median .985). Population trajectory and non-residential capacity
provide most of these cells. The observed coupling makes desirability an
important interpretive dimension of the prototype.

\begin{table}[!htbp]
\centering
\begingroup
\small
\setlength{\tabcolsep}{3.4pt}
\renewcommand{\arraystretch}{1.03}
\begin{tabular}{@{}lrrr@{}}
\toprule
\textbf{Model} & \textbf{Median} & \textbf{IQR} & \textbf{Range} \\
\midrule
Gemma2 & 11.1 & 5.8--16.5 & 1.8--38.6 \\
Mistral & 11.2 & 5.8--17.2 & 1.7--28.3 \\
Qwen2.5 & 11.3 & 8.1--16.1 & 2.8--36.9 \\
Phi-4 & 11.5 & 7.1--20.0 & 2.6--29.5 \\
Qwen3 & 11.8 & 8.9--16.5 & 2.6--26.0 \\
Llama3.1 & 12.6 & 8.6--18.3 & 4.1--36.1 \\
Gemma3 & 13.6 & 7.9--19.3 & 3.3--43.7 \\
OLMo2 & 15.5 & 10.3--19.4 & 2.7--31.7 \\
GLM-4 & 16.1 & 9.0--21.4 & 2.9--44.4 \\
DeepSeek-R1 & 23.7 & 20.5--28.6 & 11.3--45.2 \\
\bottomrule
\end{tabular}
\endgroup

\caption{Checkpoint heterogeneity across all 40 dimensions. Each checkpoint
is compared with a dependency-balanced consensus that excludes its own
component; entries are dimension-level absolute profile-rank differences in
percentage points.}
\label{tab:model-divergence}
\end{table}

  \begin{table}[!htbp]
  \centering
  \small
  \begin{tabular}{@{}lrrr@{}}
\toprule
Analysis & $n$ & Rank $\rho$ & Shift (pp) \\
\midrule
Upper tail: 30\% & 155 & 0.899 & 4.7 \\
Upper tail: 10\% & 155 & 0.908 & 5.0 \\
Checkpoint median & 28 & 0.990 & 0.0 \\
Component mean & 28 & 0.943 & 0.1 \\
Leave one component out & 12 & 0.965 & 0.5 \\
Chinese instructions & 48 & 0.950 & 4.2 \\
\bottomrule
\end{tabular}

  \caption{Sensitivity of the standardized 40-dimensional portrait. Rank
  $\rho$ compares each alternative with the primary analysis. Shift is the
  median absolute change in source-percentile rank; the leave-one-component
  row reports the least favorable correlation and largest shift across the
  seven omissions.}
  \label{tab:sensitivity}
  \end{table}
  \FloatBarrier

\FloatBarrier

\section{Discussion}

The main result is a recurring physical portrait: larger built extent, faster
population growth, greater mapped infrastructure presence, and less sparse
form. Identification varies sharply across the panel. Density, vertical form,
building age, and volume intensity provide limited cross-model evidence;
built extent, growth, infrastructure, and non-residential capacity are widely
measurable.

The 40-dimensional matrix preserves source semantics. Its shared 2,000-city
target supports comparison across 36 rows; Streetscapes and Building Atlas add
source-specific visual and geometric measures. Human/SDG and identity rows
remain contextual or normative, with row-level metadata defining their scope.

City-count, population-weighted, and region-balanced targets respectively
describe centres, resident exposure, and equal regional influence. Their broad
agreement identifies a common core, while remaining changes show that the
reference population matters.

Row-wise contrasts and maps connect the prototype to observed profiles, and
joint-profile judgments link marginal ratings to integrated city cards.
Regional tilt is substantially associated with scale and development
composition, leaving a smaller positive residual for Europe. Typicality and
desirability are strongly coupled where both are reliably measured, consistent
with culturally variable model-based urban perception
\citep{zhao2026urbanperception}.

\section{Limitations and Ethics}

The audit covers 7--14B instruction-following checkpoints and English field
descriptions; a Chinese rerun is a language sensitivity check. Urban units are
morphological centres. Remote-sensing, street-view, and map errors may be
geographically patterned, and mapped proxies only partially represent social
and environmental conditions. The joint-profile analysis spans six regions;
regional adjustment is descriptive and HDI is a broad development proxy.
Because typicality can naturalize planning norms
\citep{bender2021parrots,weidinger2022taxonomy}, the scores are intended for
research, with human and local evidence required for planning applications.

\section{Conclusion}

Across ten checkpoints and 40 dimensions, Default City exhibits a recurring
scale--growth--infrastructure--form core with regional and checkpoint
variation. Replication and direct joint-profile ratings support its stability
and behavioral coherence.

\bibliography{references}

@article{rosch1975categories,
  author = {Rosch, Eleanor},
  title = {Cognitive Representations of Semantic Categories},
  journal = {Journal of Experimental Psychology: General},
  volume = {104},
  number = {3},
  pages = {192--233},
  year = {1975},
  doi = {10.1037/0096-3445.104.3.192}
}

@article{hampton2012thinking,
  author = {Hampton, James A.},
  title = {Thinking Intuitively: The Rich (and at Times Illogical) World of Concepts},
  journal = {Current Directions in Psychological Science},
  volume = {21},
  number = {6},
  pages = {398--402},
  year = {2012},
  doi = {10.1177/0963721412457364}
}

@article{medin1978context,
  author = {Medin, Douglas L. and Schaffer, Marguerite M.},
  title = {Context Theory of Classification Learning},
  journal = {Psychological Review},
  volume = {85},
  number = {3},
  pages = {207--238},
  year = {1978},
  doi = {10.1037/0033-295X.85.3.207}
}

@article{malaviya2025contextualized,
  author = {Malaviya, Chaitanya and Chang, Joseph Chee and Roth, Dan and Iyyer, Mohit and Yatskar, Mark and Lo, Kyle},
  title = {Contextualized Evaluations: Judging Language Model Responses to Underspecified Queries},
  journal = {Transactions of the Association for Computational Linguistics},
  volume = {13},
  pages = {878--900},
  year = {2025},
  doi = {10.1162/tacl.a.24}
}

@inproceedings{santurkar2023opinions,
  author = {Santurkar, Shibani and Durmus, Esin and Ladhak, Faisal and Lee, Cinoo and Liang, Percy and Hashimoto, Tatsunori},
  title = {Whose Opinions Do Language Models Reflect?},
  booktitle = {Proceedings of the 40th International Conference on Machine Learning},
  series = {Proceedings of Machine Learning Research},
  volume = {202},
  pages = {29971--30004},
  publisher = {PMLR},
  year = {2023},
  url = {https://proceedings.mlr.press/v202/santurkar23a.html}
}

@inproceedings{ribeiro2020checklist,
  author = {Ribeiro, Marco Tulio and Wu, Tongshuang and Guestrin, Carlos and Singh, Sameer},
  title = {Beyond Accuracy: Behavioral Testing of {NLP} Models with {C}heck{L}ist},
  booktitle = {Proceedings of the 58th Annual Meeting of the Association for Computational Linguistics},
  pages = {4902--4912},
  publisher = {Association for Computational Linguistics},
  year = {2020},
  doi = {10.18653/v1/2020.acl-main.442}
}

@inproceedings{wang2024fair,
  author = {Wang, Peiyi and Li, Lei and Chen, Liang and Cai, Zefan and Zhu, Dawei and Lin, Binghuai and Cao, Yunbo and Kong, Lingpeng and Liu, Qi and Liu, Tianyu and Sui, Zhifang},
  title = {Large Language Models Are Not Fair Evaluators},
  booktitle = {Proceedings of the 62nd Annual Meeting of the Association for Computational Linguistics (Volume 1: Long Papers)},
  pages = {9440--9450},
  publisher = {Association for Computational Linguistics},
  year = {2024},
  doi = {10.18653/v1/2024.acl-long.511}
}

@inproceedings{pezeshkpour2024order,
  author = {Pezeshkpour, Pouya and Hruschka, Estevam},
  title = {Large Language Models Sensitivity to the Order of Options in Multiple-Choice Questions},
  booktitle = {Findings of the Association for Computational Linguistics: NAACL 2024},
  pages = {2006--2017},
  publisher = {Association for Computational Linguistics},
  year = {2024},
  doi = {10.18653/v1/2024.findings-naacl.130}
}

@article{manvi2024geographic,
  author = {Manvi, Rohin and Khanna, Samar and Burke, Marshall and Lobell, David and Ermon, Stefano},
  title = {Large Language Models Are Geographically Biased},
  journal = {arXiv preprint arXiv:2402.02680},
  year = {2024}
}

@inproceedings{faisal2023geographic,
  author = {Faisal, Fahim and Anastasopoulos, Antonios},
  title = {Geographic and Geopolitical Biases of Language Models},
  booktitle = {Proceedings of the 3rd Workshop on Multi-lingual Representation Learning},
  pages = {139--163},
  publisher = {Association for Computational Linguistics},
  year = {2023},
  doi = {10.18653/v1/2023.mrl-1.12}
}

@inproceedings{dunn2024populations,
  author = {Dunn, Jonathan and Adams, Benjamin and Tayyar Madabushi, Harish},
  title = {Pre-Trained Language Models Represent Some Geographic Populations Better than Others},
  booktitle = {Proceedings of the 2024 Joint International Conference on Computational Linguistics, Language Resources and Evaluation},
  pages = {12966--12976},
  publisher = {ELRA and ICCL},
  year = {2024},
  url = {https://aclanthology.org/2024.lrec-main.1135/}
}

@inproceedings{li2024land,
  author = {Li, Bryan and Haider, Samar and Callison-Burch, Chris},
  title = {This Land is {Your, My} Land: Evaluating Geopolitical Bias in Language Models through Territorial Disputes},
  booktitle = {Proceedings of the 2024 Conference of the North American Chapter of the Association for Computational Linguistics: Human Language Technologies},
  pages = {3855--3871},
  publisher = {Association for Computational Linguistics},
  year = {2024},
  doi = {10.18653/v1/2024.naacl-long.213}
}

@article{liang2023helm,
  author = {Liang, Percy and Bommasani, Rishi and Lee, Tony and Tsipras, Dimitris and Soylu, Dilara and Yasunaga, Michihiro and Zhang, Yian and Narayanan, Deepak and Wu, Yuhuai and Kumar, Ananya and others},
  title = {Holistic Evaluation of Language Models},
  journal = {Transactions on Machine Learning Research},
  year = {2023},
  url = {https://openreview.net/forum?id=iO4LZibEqW}
}

@article{zhao2026urbanperception,
  author = {Zhao, Rong and Liu, Wanqi and Sha, Zhizhou and Su, Nanxi and Zhang, Yecheng and Long, Ying},
  title = {Culturally Uneven Urban Perception in Large Language Models},
  journal = {arXiv preprint arXiv:2604.20048},
  year = {2026},
  doi = {10.48550/arXiv.2604.20048},
  url = {https://arxiv.org/abs/2604.20048}
}

@inproceedings{mitchell2019modelcards,
  author = {Mitchell, Margaret and Wu, Simone and Zaldivar, Andrew and Barnes, Parker and Vasserman, Lucy and Hutchinson, Ben and Spitzer, Elena and Raji, Inioluwa Deborah and Gebru, Timnit},
  title = {Model Cards for Model Reporting},
  booktitle = {Proceedings of the Conference on Fairness, Accountability, and Transparency},
  pages = {220--229},
  publisher = {Association for Computing Machinery},
  year = {2019},
  doi = {10.1145/3287560.3287596}
}

@article{gebru2021datasheets,
  author = {Gebru, Timnit and Morgenstern, Jamie and Vecchione, Briana and Vaughan, Jennifer Wortman and Wallach, Hanna and Daum\'{e}, Hal and Crawford, Kate},
  title = {Datasheets for Datasets},
  journal = {Communications of the ACM},
  volume = {64},
  number = {12},
  pages = {86--92},
  year = {2021},
  doi = {10.1145/3458723}
}

@article{pineau2021reproducibility,
  author = {Pineau, Joelle and Vincent-Lamarre, Philippe and Sinha, Koustuv and Larivi\`ere, Vincent and Beygelzimer, Alina and d'Alch\'{e}-Buc, Florence and Fox, Emily and Larochelle, Hugo},
  title = {Improving Reproducibility in Machine Learning Research: A Report from the {NeurIPS} 2019 Reproducibility Program},
  journal = {Journal of Machine Learning Research},
  volume = {22},
  number = {164},
  pages = {1--20},
  year = {2021},
  url = {https://www.jmlr.org/papers/v22/20-303.html}
}

@inproceedings{pattnayak2026reproevalcard,
  author = {Pattnayak, Priyaranjan and Bhatia, Apoorv},
  title = {{ReproEvalCard}: A Reporting Standard for Reproducible Evaluation of {LLM} Pipelines},
  booktitle = {Proceedings of the 64th Annual Meeting of the Association for Computational Linguistics (Volume 2: Short Papers)},
  pages = {238--249},
  publisher = {Association for Computational Linguistics},
  year = {2026},
  doi = {10.18653/v1/2026.acl-short.22}
}

@inproceedings{bender2021parrots,
  author = {Bender, Emily M. and Gebru, Timnit and McMillan-Major, Angelina and Shmitchell, Shmargaret},
  title = {On the Dangers of Stochastic Parrots: Can Language Models Be Too Big?},
  booktitle = {Proceedings of the 2021 ACM Conference on Fairness, Accountability, and Transparency},
  pages = {610--623},
  publisher = {Association for Computing Machinery},
  year = {2021},
  doi = {10.1145/3442188.3445922}
}

@inproceedings{weidinger2022taxonomy,
  author = {Weidinger, Laura and Uesato, Jonathan and Rauh, Maribeth and Griffin, Conor and Huang, Po-Sen and Mellor, John and Glaese, Amelia and Cheng, Myra and Balle, Borja and Kasirzadeh, Atoosa and others},
  title = {Taxonomy of Risks Posed by Language Models},
  booktitle = {Proceedings of the 2022 ACM Conference on Fairness, Accountability, and Transparency},
  pages = {214--229},
  publisher = {Association for Computing Machinery},
  year = {2022},
  doi = {10.1145/3531146.3533088}
}

@techreport{melchiorri2024stats,
  author = {Melchiorri, Michele and Mari Rivero, Ines and Florio, Pietro and Schiavina, Marcello and Krasnodebska, Katarzyna and Politis, Panagiotis and Uhl, Johannes and Pesaresi, Martino and others},
  title = {Stats in the City: The {GHSL} Urban Centre Database 2025},
  institution = {Publications Office of the European Union},
  number = {JRC139768},
  year = {2024},
  doi = {10.2760/3046391}
}

@article{dijkstra2021degree,
  author = {Dijkstra, Lewis and Florczyk, Aneta J. and Freire, Sergio and Kemper, Thomas and Melchiorri, Michele and Pesaresi, Martino and Schiavina, Marcello},
  title = {Applying the Degree of Urbanisation to the Globe: A New Harmonised Definition Reveals a Different Picture of Global Urbanisation},
  journal = {Journal of Urban Economics},
  volume = {125},
  pages = {103312},
  year = {2021},
  doi = {10.1016/j.jue.2020.103312}
}

@article{rao1992resampling,
  author = {Rao, J. N. K. and Wu, C. F. J. and Yue, K.},
  title = {Some Recent Work on Resampling Methods for Complex Surveys},
  journal = {Survey Methodology},
  volume = {18},
  number = {2},
  pages = {209--217},
  year = {1992}
}

@article{beaumont2012bootstrap,
  author = {Beaumont, Jean-Fran\c{c}ois and Patak, Zdenek},
  title = {On the Generalized Bootstrap for Sample Surveys with Special Attention to Poisson Sampling},
  journal = {International Statistical Review},
  volume = {80},
  number = {1},
  pages = {127--148},
  year = {2012},
  doi = {10.1111/j.1751-5823.2011.00166.x}
}

@article{feng2024citybench,
  author = {Feng, Jie and Zhang, Jun and Liu, Tianhui and Zhang, Xin and Ouyang, Tianjian and Yan, Junbo and Du, Yuwei and Guo, Siqi and Li, Yong},
  title = {{CityBench}: Evaluating the Capabilities of Large Language Models for Urban Tasks},
  journal = {arXiv preprint arXiv:2406.13945},
  year = {2024}
}

@article{zheng2025urbanplanbench,
  author = {Zheng, Yu and Liu, Longyi and Lin, Yuming and Feng, Jie and Zhang, Guozhen and Jin, Depeng and Li, Yong},
  title = {{UrbanPlanBench}: A Comprehensive Urban Planning Benchmark for Evaluating Large Language Models},
  journal = {arXiv preprint arXiv:2504.21027},
  year = {2025}
}

@article{zhang2025ai4us,
  author = {Zhang, Yecheng and Zhao, Rong and Huang, Zimu and Wang, Xinyu and Ma, Yue and Long, Ying},
  title = {{GenAI} Models Capture Urban Science but Oversimplify Complexity},
  journal = {arXiv preprint arXiv:2505.13803},
  year = {2025},
  doi = {10.48550/arXiv.2505.13803},
  url = {https://arxiv.org/abs/2505.13803}
}

@inproceedings{campanella2024bigcity,
  author = {Campanella, Charlie and van der Goot, Rob},
  title = {Big City Bias: Evaluating the Impact of Metropolitan Size on Computational Job Market Abilities of Language Models},
  booktitle = {Proceedings of the First Workshop on Natural Language Processing for Human Resources},
  pages = {73--77},
  publisher = {Association for Computational Linguistics},
  year = {2024},
  doi = {10.18653/v1/2024.nlp4hr-1.6},
  url = {https://aclanthology.org/2024.nlp4hr-1.6/}
}

@article{yang2025qwen3,
  author = {Yang, An and Li, Anfeng and Yang, Baosong and Zhang, Beichen and Hui, Binyuan and Zheng, Bo and others},
  title = {{Qwen3} Technical Report},
  journal = {arXiv preprint arXiv:2505.09388},
  year = {2025}
}

@article{olmo2025furious,
  author = {{OLMo Team}},
  title = {2 {OLMo} 2 Furious},
  journal = {arXiv preprint arXiv:2501.00656},
  year = {2025}
}

@misc{mistral2024nemo,
  author = {{Mistral AI Team}},
  title = {Mistral {NeMo}},
  year = {2024},
  howpublished = {Model release and documentation},
  url = {https://mistral.ai/news/mistral-nemo/}
}

@article{gemmateam2024gemma2,
  title = {Gemma 2: Improving Open Language Models at a Practical Size},
  author = {{Gemma Team}},
  journal = {arXiv preprint arXiv:2408.00118},
  year = {2024}
}

@article{dubey2024llama3,
  title = {The Llama 3 Herd of Models},
  author = {Dubey, Abhimanyu and others},
  journal = {arXiv preprint arXiv:2407.21783},
  year = {2024}
}

@article{hou2024globalstreetscapes,
  author = {Hou, Yujun and Quintana, Matias and Khomiakov, Maxim and Yap, Winston and Ouyang, Jiani and Ito, Koichi and Wang, Zeyu and Zhao, Tianhong and Biljecki, Filip},
  title = {Global Streetscapes: A Comprehensive Dataset of 10 Million Street-Level Images across 688 Cities for Urban Science and Analytics},
  journal = {ISPRS Journal of Photogrammetry and Remote Sensing},
  volume = {215},
  pages = {216--238},
  year = {2024},
  doi = {10.1016/j.isprsjprs.2024.06.023}
}

@article{zhu2025globalbuildingatlas,
  author = {Zhu, Xiao Xiang and Chen, Sining and Zhang, Fahong and Shi, Yilei and Wang, Yuanyuan},
  title = {{GlobalBuildingAtlas}: An Open Global and Complete Dataset of Building Polygons, Heights and {LoD1} 3D Models},
  journal = {arXiv preprint arXiv:2506.04106},
  year = {2025}
}

@misc{bondarenko2025worldpop,
  author = {Bondarenko, Maksym and Priyatikanto, Rhorom and Tejedor-Garavito, Natalia and Zhang, Wei and McKeen, Taylor and Cunningham, Andrew and Woods, Thomas and Hilton, Jason and Cihan, Doga and Nosatiuk, Bohdan and Brinkhoff, Thomas and Tatem, Andrew and Sorichetta, Alessandro},
  title = {The Spatial Distribution of Population Broken Down by Gender and Age Groupings in 2015--2030 at 30 Arc-Second Resolution, {R2025A} v1},
  howpublished = {WorldPop, University of Southampton},
  year = {2025},
  doi = {10.5258/SOTON/WP00846}
}

@misc{overture2026data,
  author = {{Overture Maps Foundation}},
  title = {Overture Maps Data Documentation: Places, Buildings, and Transportation},
  howpublished = {\url{https://docs.overturemaps.org/}},
  year = {2026}
}

@misc{mobilitydata2026gtfs,
  author = {{MobilityData}},
  title = {General Transit Feed Specification},
  howpublished = {\url{https://gtfs.org/}},
  year = {2026}
}
  \clearpage
  \appendix
  \setcounter{secnumdepth}{1}%
  \setcounter{figure}{0}%
  \setcounter{table}{0}%
  \renewcommand{\thefigure}{S\arabic{figure}}%
  \renewcommand{\thetable}{S\arabic{table}}%
  \renewcommand{\theHfigure}{supp.figure.\arabic{figure}}%
  \renewcommand{\theHtable}{supp.table.\arabic{table}}%
  \def\ArxivIntegratedAppendix{1}%
  The 40 dimensions form one experiment and reporting matrix. Reliability,
ordering, matched-construct, and replication metadata remain attached to the
relevant rows and cells. Thirty-six rows use the common 2,000-city target,
three use the complete 356-city Global Streetscapes frame, and one uses the
353-city GBA--Streetscapes joint frame.

Every display follows the same seven-domain order:
\emph{Scale, Extent \& Change}; \emph{Built Form \& Land Use};
\emph{Networks, Mobility \& Infrastructure}; \emph{Ecology \& Environment};
\emph{Population, Housing \& Public Life};
\emph{Economy, Function \& Centrality}; and
\emph{Culture, Institutions \& Identity}. All associated contextual and
source-specific fields are reported.

\ifdefined\ArxivIntegratedAppendix
\begin{table}[!ht]
\centering
\footnotesize
\setlength{\tabcolsep}{4pt}
\begin{tabular}{@{}p{.25\columnwidth}p{.65\columnwidth}@{}}
\toprule
Appendix item & Added information \\
\midrule
Tables~\ref{tab:dimension-ledger} and \ref{tab:source-ledger} & Complete
measurement status, source roles, and target frames \\
Tables~\ref{tab:parameter-a}--\ref{tab:parameter-b} & All 40 parameters,
400 checkpoint-specific Defaults, combined values, and percentiles \\
Figures~\ref{fig:maps-a}--\ref{fig:maps-d} & Complete global views with
reliable-checkpoint counts and source-frame coverage \\
Figure~\ref{fig:regional-expanded} and Table~\ref{tab:regional-grids} &
Complete domain--region and checkpoint--region values \\
Tables~\ref{tab:divergence-a}--\ref{tab:divergence-b} & All 400
dimension--checkpoint leave-component-out divergences \\
Figures~\ref{fig:status-a}--\ref{fig:status-b} & Every clean,
low-dispersion, and order/reduction-sensitive measurement cell \\
Table~\ref{tab:joint-profile} & Checkpoint-level marginal--joint and
forward--reverse correlations with bootstrap intervals \\
Table~\ref{tab:regional-controls} & Population, urban-scale, HDI, and joint
regional conditioning variants \\
Tables~\ref{tab:model-roster}--\ref{tab:runtime-receipts} & Prompt text,
checkpoint pins, inference settings, and artifact receipts \\
\bottomrule
\end{tabular}
\caption{Roadmap to the integrated appendix. The sensitivity summary is
promoted to the main text; measurement and source audits remain here with the
complete results.}
\label{tab:appendix-roadmap}
\end{table}
\else
\begin{table}[!ht]
\centering
\footnotesize
\setlength{\tabcolsep}{3pt}
\begin{tabular}{@{}p{.22\columnwidth}p{.69\columnwidth}@{}}
\toprule
Main item & Appendix extension and added information \\
\midrule
Table 1 & Tables~\ref{tab:parameter-a}--\ref{tab:parameter-b}: all 40
parameters, 400 model Defaults, combined values, and percentiles \\
Figure 2 & Figures~\ref{fig:maps-a}--\ref{fig:maps-d}: all 40 global views,
reliable-checkpoint counts, and source-frame coverage \\
Figure 3 & Figure~\ref{fig:regional-expanded} and
Tables~\ref{tab:regional-grids} and \ref{tab:regional-controls}: every
domain--region and checkpoint--region value, plus all conditioning variants \\
Table 2 & Tables~\ref{tab:divergence-a}--\ref{tab:divergence-b}: every
dimension--checkpoint leave-component-out divergence \\
Figure 4 & Table~\ref{tab:joint-profile}: all ten marginal--joint and
forward--reverse correlations with city-bootstrap intervals \\
Sensitivity & Table~\ref{tab:sensitivity}: threshold, aggregation,
leave-one-component, and Chinese-instruction checks \\
Measurement & Figures~\ref{fig:status-a}--\ref{fig:status-b}: all 400
clean, low-dispersion, and order/reduction-sensitive cells \\
\bottomrule
\end{tabular}
\caption{Reading guide. The appendix extends the main displays on one
40-dimensional reporting surface.}
\label{tab:reading-guide}
\end{table}
\fi

\section{Study Inventory and Parameter Results}

\begin{table}[!htbp]
\centering
\begingroup
\small
\renewcommand{\arraystretch}{0.92}
\begin{tabular*}{\textwidth}{@{\extracolsep{\fill}}r l r r r r@{}}
\toprule
G & Dimension & Frame & C & L & O \\
\midrule
1 & Population & 2,000 & 8 & 2 & 0 \\
1 & Density & 2,000 & 1 & 9 & 0 \\
1 & Built extent & 2,000 & 6 & 4 & 0 \\
1 & Footprint+ & 2,000 & 6 & 4 & 0 \\
1 & Pop. growth & 2,000 & 8 & 2 & 0 \\
1 & Expand/dens. & 2,000 & 8 & 2 & 0 \\
1 & Land-use eff. & 2,000 & 2 & 0 & 8 \\
\addlinespace[1.2pt]
2 & Vertical & 2,000 & 1 & 5 & 4 \\
2 & Volume & 2,000 & 0 & 10 & 0 \\
2 & Bldg. type & 2,000 & 3 & 1 & 6 \\
2 & Bldg. age & 2,000 & 0 & 2 & 8 \\
2 & LCZ* & 2,000$^\ast$ & 5 & 1 & 4 \\
2 & Land cover* & 2,000$^\ast$ & 7 & 1 & 2 \\
2 & Building grain & 2,000 & 0 & 0 & 10 \\
2 & Street view & 356 & 1 & 0 & 9 \\
2 & Visual enclosure & 356 & 4 & 0 & 6 \\
2 & GBA enclosure & 353$^\ast$ & 3 & 7 & 0 \\
\addlinespace[1.2pt]
3 & Roads & 2,000 & 2 & 5 & 3 \\
3 & Infrastructure & 2,000 & 9 & 1 & 0 \\
3 & Digital access & 2,000$^\ast$ & 3 & 0 & 7 \\
\addlinespace[1.2pt]
3 & Air/port nodes & 2,000 & 0 & 0 & 10 \\
3 & GTFS presence & 2,000 & 1 & 0 & 9 \\
3 & Street topology & 2,000 & 0 & 0 & 10 \\
\addlinespace[1.2pt]
4 & Greenness & 2,000 & 5 & 2 & 3 \\
4 & Canopy & 2,000 & 9 & 1 & 0 \\
4 & Climate & 2,000 & 3 & 0 & 7 \\
4 & Emissions & 2,000 & 0 & 0 & 10 \\
4 & Green access & 2,000 & 9 & 1 & 0 \\
4 & Visible veg. & 356 & 8 & 2 & 0 \\
\addlinespace[1.2pt]
5 & Age--sex & 2,000$^\ast$ & 0 & 0 & 10 \\
5 & Subnational HDI & 2,000$^\ast$ & 8 & 2 & 0 \\
5 & Mapped health & 2,000$^\ast$ & 2 & 0 & 8 \\
5 & Everyday svc. & 2,000$^\ast$ & 9 & 1 & 0 \\
\addlinespace[1.2pt]
6 & Non-res. cap.+ & 2,000 & 9 & 1 & 0 \\
6 & Economic output & 2,000 & 0 & 0 & 10 \\
6 & Functional extent & 2,000$^\ast$ & 7 & 0 & 3 \\
6 & POI diversity & 2,000$^\ast$ & 0 & 0 & 10 \\
\addlinespace[1.2pt]
7 & Capital role & 2,000 & 6 & 4 & 0 \\
7 & Heritage & 2,000 & 0 & 1 & 9 \\
7 & Civic POIs & 2,000$^\ast$ & 9 & 1 & 0 \\
\bottomrule
\end{tabular*}
\endgroup

\caption{Master 40-dimensional design ledger and row order used in
every subsequent display. G1--G7 follow the seven-domain order stated on the
first page. Frame is the row-specific target frame; an asterisk marks
incomplete measurement within that target. C, L, and O count clean,
low-dispersion, and ordering/reduction-sensitive checkpoint cells.}
\label{tab:dimension-ledger}
\end{table}

\begin{table}[!htbp]
\centering
\small
\setlength{\tabcolsep}{3pt}
\renewcommand{\arraystretch}{1.05}
\begin{tabular}{@{}p{.14\textwidth}rp{.19\textwidth}p{.27\textwidth}p{.26\textwidth}@{}}
\toprule
Source family & Products & Frame role & Direct role & Interpretation \\
\midrule
JRC GHSL & 2 & Common 2,000-city target and replication & GHS-UCDB themes
and partial GHS-FUA crosswalk & Morphological centres; modeled fields retain
their source semantics \\
UNESCO & 1 & Common target & Heritage representative-point count and
distance & Proxy for designated heritage presence \\
OurAirports & 1 & Common target & Scheduled-airport nodes & Node presence \\
NGA & 1 & Common target & Medium/large-port nodes & Node presence \\
MobilityData & 1 & Common target & Active official GTFS catalog bounding
boxes & Catalog coverage indicator \\
Overture Maps & 3 & Common target & Places, Buildings, and Transportation
themes & Mapped composition and morphology \\
Global Streetscapes & 1 & Complete 356-city availability frame & Physical
appearance and visual-enclosure fields & Source-conditioned image
availability \\
Global Building Atlas & 3 & Complete 353-city joint frame & LoD1 polygons,
height, and enclosure geometry & Joint-availability parameter; raw building
payloads excluded \\
\bottomrule
\end{tabular}
\caption{Thirteen directly materialized products in eight source families.
WorldPop-derived and other thematic lineages inherited through GHS-UCDB
remain recorded as upstream provenance.}
\label{tab:source-ledger}
\end{table}

\FloatBarrier

\begin{table}[!htbp]
\centering
\begingroup
\footnotesize
\setlength{\tabcolsep}{0.70pt}
\renewcommand{\arraystretch}{1.04}
\begin{tabular*}{\textwidth}{@{\extracolsep{\fill}}>{\raggedright\arraybackslash}p{.215\textwidth}r*{10}{r}rr@{}}
\toprule
\textbf{Parameter} & \textbf{Global} & \multicolumn{10}{c}{\textbf{Model-specific Default}} & \textbf{Combined} & \textbf{Pct.} \\
\cmidrule(lr){3-12}
 &  & \multicolumn{1}{c}{Q3} & \multicolumn{1}{c}{Q2.5} & \multicolumn{1}{c}{DS} & \multicolumn{1}{c}{G2} & \multicolumn{1}{c}{G3} & \multicolumn{1}{c}{L3.1} & \multicolumn{1}{c}{MN} & \multicolumn{1}{c}{O2} & \multicolumn{1}{c}{P4} & \multicolumn{1}{c}{GLM} &  &  \\
\midrule
\multicolumn{14}{@{}l}{\textbf{Scale, Extent \& Change}} \\
\rule{0pt}{2.35ex}Population (millions) & 0.11 & 0.50 & 0.50 & 0.07 & 0.50 & 0.28 & 0.49 & \cellcolor[gray]{0.90}0.50 & \cellcolor[gray]{0.90}0.47 & 0.50 & 0.50 & \textbf{0.50} & 90 \\
\rule{0pt}{2.35ex}Density (people/km$^2$) & 4774 & \cellcolor[gray]{0.90}5079 & 9175 & \cellcolor[gray]{0.90}6892 & \cellcolor[gray]{0.90}7388 & \cellcolor[gray]{0.90}9634 & \cellcolor[gray]{0.90}5905 & \cellcolor[gray]{0.90}9371 & \cellcolor[gray]{0.90}8698 & \cellcolor[gray]{0.90}9051 & \cellcolor[gray]{0.90}6594 & \textbf{9175} & 88 \\
\rule{0pt}{2.35ex}Built surface (km$^2$) & 4.5 & 21.2 & 21.6 & \cellcolor[gray]{0.90}7.6 & 20.8 & 21.2 & \cellcolor[gray]{0.90}18.8 & 19.5 & \cellcolor[gray]{0.90}16.5 & 21.0 & \cellcolor[gray]{0.90}21.0 & \textbf{21.0} & 90 \\
\rule{0pt}{2.35ex}Footprint compactness (0--1) & 0.42 & 0.54 & 0.54 & \cellcolor[gray]{0.90}0.49 & 0.54 & 0.55 & \cellcolor[gray]{0.90}0.54 & \cellcolor[gray]{0.90}0.54 & 0.54 & 0.54 & \cellcolor[gray]{0.90}0.54 & \textbf{0.54} & 86 \\
\rule{0pt}{2.35ex}Population growth (\%/yr) & 0.3 & 1.7 & 2.2 & \cellcolor[gray]{0.90}0.5 & 2.2 & 2.2 & 2.2 & 2.2 & 2.2 & 2.2 & \cellcolor[gray]{0.90}-0.7 & \textbf{2.2} & 90 \\
\rule{0pt}{2.35ex}Pop.--built growth gap (pp/yr) & -0.4 & 1.1 & 1.4 & \cellcolor[gray]{0.90}0.4 & 1.1 & 1.3 & 1.2 & 1.3 & 0.8 & 1.4 & \cellcolor[gray]{0.90}1.4 & \textbf{1.2} & 87 \\
\rule{0pt}{2.35ex}Land/population growth ratio & 0.68 & \cellcolor[gray]{0.90}0.92 & \cellcolor[gray]{0.90}0.94 & \cellcolor[gray]{0.90}1.46 & 0.88 & \cellcolor[gray]{0.90}0.88 & \cellcolor[gray]{0.90}0.82 & 0.94 & \cellcolor[gray]{0.90}0.73 & \cellcolor[gray]{0.90}0.80 & \cellcolor[gray]{0.90}1.04 & \textbf{0.91} & 59 \\
\midrule
\multicolumn{14}{@{}l}{\textbf{Built Form \& Land Use}} \\
\rule{0pt}{2.35ex}Mean built height (m) & 6.6 & \cellcolor[gray]{0.90}8.0 & \cellcolor[gray]{0.90}7.0 & \cellcolor[gray]{0.90}7.2 & \cellcolor[gray]{0.90}8.1 & \cellcolor[gray]{0.90}8.0 & \cellcolor[gray]{0.90}7.5 & \cellcolor[gray]{0.90}8.8 & \cellcolor[gray]{0.90}8.3 & 8.6 & \cellcolor[gray]{0.90}8.4 & \textbf{8.6} & 81 \\
\rule{0pt}{2.35ex}Built volume density ($10^6$m$^3$/km$^2$) & 1.59 & \cellcolor[gray]{0.90}2.11 & \cellcolor[gray]{0.90}2.06 & \cellcolor[gray]{0.90}0.98 & \cellcolor[gray]{0.90}1.82 & \cellcolor[gray]{0.90}1.94 & \cellcolor[gray]{0.90}1.88 & \cellcolor[gray]{0.90}1.94 & \cellcolor[gray]{0.90}1.78 & \cellcolor[gray]{0.90}1.71 & \cellcolor[gray]{0.90}1.75 & \textbf{--} & -- \\
\rule{0pt}{2.35ex}Residential stock share (\%) & 97.4 & \cellcolor[gray]{0.90}86.1 & \cellcolor[gray]{0.90}97.2 & \cellcolor[gray]{0.90}84.5 & 83.7 & 82.7 & \cellcolor[gray]{0.90}100.0 & \cellcolor[gray]{0.90}100.0 & 93.8 & \cellcolor[gray]{0.90}91.8 & \cellcolor[gray]{0.90}99.0 & \textbf{88.5} & 27 \\
\rule{0pt}{2.35ex}Median built-stock epoch & 1990 & \cellcolor[gray]{0.90}1995 & \cellcolor[gray]{0.90}1995 & \cellcolor[gray]{0.90}2002 & \cellcolor[gray]{0.90}2000 & \cellcolor[gray]{0.90}1975 & \cellcolor[gray]{0.90}2000 & \cellcolor[gray]{0.90}2000 & \cellcolor[gray]{0.90}1992 & \cellcolor[gray]{0.90}2005 & \cellcolor[gray]{0.90}1975 & \textbf{--} & -- \\
\rule{0pt}{2.35ex}Compact LCZ share (\%) & 5.6 & \cellcolor[gray]{0.90}14.8 & \cellcolor[gray]{0.90}7.9 & \cellcolor[gray]{0.90}7.9 & 17.5 & \cellcolor[gray]{0.90}18.7 & 13.5 & 9.2 & \cellcolor[gray]{0.90}12.7 & 2.9 & 4.6 & \textbf{9.2} & 62 \\
\rule{0pt}{2.35ex}Built land share (\%) & 28.3 & 76.3 & 78.4 & \cellcolor[gray]{0.90}60.0 & 68.8 & 83.5 & 83.2 & \cellcolor[gray]{0.90}83.0 & \cellcolor[gray]{0.90}59.9 & 79.1 & 83.9 & \textbf{79.1} & 86 \\
\rule{0pt}{2.35ex}Median building footprint (m$^2$) & 80 & \cellcolor[gray]{0.90}87 & \cellcolor[gray]{0.90}77 & \cellcolor[gray]{0.90}84 & \cellcolor[gray]{0.90}88 & \cellcolor[gray]{0.90}113 & \cellcolor[gray]{0.90}81 & \cellcolor[gray]{0.90}82 & \cellcolor[gray]{0.90}69 & \cellcolor[gray]{0.90}61 & \cellcolor[gray]{0.90}76 & \textbf{--} & -- \\
\rule{0pt}{2.35ex}Visible building share (\%) & 14.5 & \cellcolor[gray]{0.90}24.8 & \cellcolor[gray]{0.90}22.5 & \cellcolor[gray]{0.90}18.4 & 25.5 & \cellcolor[gray]{0.90}25.7 & \cellcolor[gray]{0.90}22.5 & \cellcolor[gray]{0.90}22.2 & \cellcolor[gray]{0.90}24.5 & \cellcolor[gray]{0.90}26.1 & \cellcolor[gray]{0.90}22.6 & \textbf{25.5} & 85 \\
\rule{0pt}{2.35ex}Visual enclosure balance (0--1) & 0.43 & \cellcolor[gray]{0.90}0.68 & 0.67 & \cellcolor[gray]{0.90}0.64 & \cellcolor[gray]{0.90}0.68 & \cellcolor[gray]{0.90}0.68 & 0.68 & \cellcolor[gray]{0.90}0.68 & 0.67 & 0.67 & \cellcolor[gray]{0.90}0.67 & \textbf{0.67} & 89 \\
\rule{0pt}{2.35ex}Height/street-width ratio & 0.29 & 0.46 & 0.43 & \cellcolor[gray]{0.90}0.34 & \cellcolor[gray]{0.90}0.46 & \cellcolor[gray]{0.90}0.18 & \cellcolor[gray]{0.90}0.46 & \cellcolor[gray]{0.90}0.46 & \cellcolor[gray]{0.90}0.36 & 0.46 & \cellcolor[gray]{0.90}0.46 & \textbf{0.45} & 89 \\
\midrule
\multicolumn{14}{@{}l}{\textbf{Networks, Mobility \& Infrastructure}} \\
\rule{0pt}{2.35ex}Road density (km/km$^2$) & 10.3 & \cellcolor[gray]{0.90}18.3 & \cellcolor[gray]{0.90}18.7 & \cellcolor[gray]{0.90}8.7 & 16.7 & \cellcolor[gray]{0.90}18.1 & \cellcolor[gray]{0.90}18.7 & \cellcolor[gray]{0.90}19.2 & \cellcolor[gray]{0.90}12.6 & 17.2 & \cellcolor[gray]{0.90}12.8 & \textbf{17.0} & 73 \\
\rule{0pt}{2.35ex}Transport-infrastructure index & 0.013 & 0.036 & 0.032 & \cellcolor[gray]{0.90}0.032 & 0.031 & 0.033 & 0.031 & 0.034 & 0.033 & 0.033 & 0.033 & \textbf{0.033} & 85 \\
\rule{0pt}{2.35ex}Fixed download speed (Mbps) & 79 & \cellcolor[gray]{0.90}238 & \cellcolor[gray]{0.90}297 & \cellcolor[gray]{0.90}153 & \cellcolor[gray]{0.90}275 & 281 & \cellcolor[gray]{0.90}262 & 275 & \cellcolor[gray]{0.90}272 & 289 & \cellcolor[gray]{0.90}209 & \textbf{281} & 84 \\
\bottomrule
\end{tabular*}
\endgroup

\caption{Expanded Table~1, part A: dimensions 1--20, preserving the main
table's column logic. Continuous entries are design-weighted medians within
each model's upper typicality quintile; GTFS coverage and national-capital
status are weighted shares. Global is the row-specific benchmark. Combined
first medians clean checkpoints within dependency components and then across
components; Pct. is its percentile in the same frame. Gray cells fall below
the measurement-reliability criterion, remain descriptive, and are excluded from
Combined.}
\label{tab:parameter-a}
\end{table}

\begin{table}[!htbp]
\centering
\begingroup
\footnotesize
\setlength{\tabcolsep}{0.70pt}
\renewcommand{\arraystretch}{1.04}
\begin{tabular*}{\textwidth}{@{\extracolsep{\fill}}>{\raggedright\arraybackslash}p{.215\textwidth}r*{10}{r}rr@{}}
\toprule
\textbf{Parameter} & \textbf{Global} & \multicolumn{10}{c}{\textbf{Model-specific Default}} & \textbf{Combined} & \textbf{Pct.} \\
\cmidrule(lr){3-12}
 &  & \multicolumn{1}{c}{Q3} & \multicolumn{1}{c}{Q2.5} & \multicolumn{1}{c}{DS} & \multicolumn{1}{c}{G2} & \multicolumn{1}{c}{G3} & \multicolumn{1}{c}{L3.1} & \multicolumn{1}{c}{MN} & \multicolumn{1}{c}{O2} & \multicolumn{1}{c}{P4} & \multicolumn{1}{c}{GLM} &  &  \\
\midrule
\multicolumn{14}{@{}l}{\textbf{Networks, Mobility \& Infrastructure}} \\
\rule{0pt}{2.35ex}Nearest scheduled airport (km) & 51 & \cellcolor[gray]{0.90}15 & \cellcolor[gray]{0.90}16 & \cellcolor[gray]{0.90}61 & \cellcolor[gray]{0.90}15 & \cellcolor[gray]{0.90}14 & \cellcolor[gray]{0.90}14 & \cellcolor[gray]{0.90}14 & \cellcolor[gray]{0.90}25 & \cellcolor[gray]{0.90}11 & \cellcolor[gray]{0.90}23 & \textbf{--} & -- \\
\rule{0pt}{2.35ex}Official GTFS coverage (\%) & 41.3 & 100.0 & \cellcolor[gray]{0.90}100.0 & \cellcolor[gray]{0.90}100.0 & \cellcolor[gray]{0.90}100.0 & \cellcolor[gray]{0.90}100.0 & \cellcolor[gray]{0.90}100.0 & \cellcolor[gray]{0.90}100.0 & \cellcolor[gray]{0.90}100.0 & \cellcolor[gray]{0.90}100.0 & \cellcolor[gray]{0.90}100.0 & \textbf{100.0} & -- \\
\rule{0pt}{2.35ex}Central road density (km/km$^2$) & 9.2 & \cellcolor[gray]{0.90}15.3 & \cellcolor[gray]{0.90}17.3 & \cellcolor[gray]{0.90}7.8 & \cellcolor[gray]{0.90}11.5 & \cellcolor[gray]{0.90}11.6 & \cellcolor[gray]{0.90}13.6 & \cellcolor[gray]{0.90}10.4 & \cellcolor[gray]{0.90}9.3 & \cellcolor[gray]{0.90}14.3 & \cellcolor[gray]{0.90}10.0 & \textbf{--} & -- \\
\midrule
\multicolumn{14}{@{}l}{\textbf{Ecology \& Environment}} \\
\rule{0pt}{2.35ex}Built-area greenness (0--1) & 0.38 & \cellcolor[gray]{0.90}0.48 & 0.51 & \cellcolor[gray]{0.90}0.44 & 0.50 & \cellcolor[gray]{0.90}0.44 & \cellcolor[gray]{0.90}0.49 & 0.52 & \cellcolor[gray]{0.90}0.49 & 0.49 & 0.51 & \textbf{0.51} & 84 \\
\rule{0pt}{2.35ex}Canopy height (m) & 8.6 & 13.2 & 13.4 & \cellcolor[gray]{0.90}7.8 & 9.6 & 13.9 & 13.3 & 13.9 & 9.7 & 12.6 & 12.4 & \textbf{12.6} & 81 \\
\rule{0pt}{2.35ex}Mean temperature ($^\circ$C) & 21.8 & \cellcolor[gray]{0.90}25.2 & \cellcolor[gray]{0.90}25.7 & \cellcolor[gray]{0.90}22.8 & \cellcolor[gray]{0.90}17.1 & 26.2 & \cellcolor[gray]{0.90}23.4 & \cellcolor[gray]{0.90}25.3 & 26.6 & \cellcolor[gray]{0.90}24.1 & 26.6 & \textbf{26.6} & 84 \\
\rule{0pt}{2.35ex}Modeled CO2 per resident & 0.53 & \cellcolor[gray]{0.90}0.73 & \cellcolor[gray]{0.90}1.01 & \cellcolor[gray]{0.90}1.49 & \cellcolor[gray]{0.90}0.22 & \cellcolor[gray]{0.90}0.18 & \cellcolor[gray]{0.90}1.04 & \cellcolor[gray]{0.90}0.40 & \cellcolor[gray]{0.90}0.40 & \cellcolor[gray]{0.90}0.34 & \cellcolor[gray]{0.90}0.86 & \textbf{--} & -- \\
\rule{0pt}{2.35ex}High-green population (\%) & 2.7 & 26.6 & 26.6 & \cellcolor[gray]{0.90}7.7 & 24.7 & 26.0 & 26.6 & 26.6 & 26.6 & 25.2 & 26.6 & \textbf{26.6} & 90 \\
\rule{0pt}{2.35ex}Visible vegetation (\%) & 14.2 & 25.9 & 25.7 & 7.9 & \cellcolor[gray]{0.90}26.0 & 9.5 & 24.3 & 26.1 & 25.9 & 25.2 & \cellcolor[gray]{0.90}24.6 & \textbf{25.5} & 88 \\
\midrule
\multicolumn{14}{@{}l}{\textbf{Population, Housing \& Public Life}} \\
\rule{0pt}{2.35ex}Young population share (\%) & 24.4 & \cellcolor[gray]{0.90}23.2 & \cellcolor[gray]{0.90}36.8 & \cellcolor[gray]{0.90}27.7 & \cellcolor[gray]{0.90}26.1 & \cellcolor[gray]{0.90}22.2 & \cellcolor[gray]{0.90}21.2 & \cellcolor[gray]{0.90}20.4 & \cellcolor[gray]{0.90}22.5 & \cellcolor[gray]{0.90}31.5 & \cellcolor[gray]{0.90}18.5 & \textbf{--} & -- \\
\rule{0pt}{2.35ex}Subnational HDI (0--1) & 0.71 & 0.87 & 0.87 & \cellcolor[gray]{0.90}0.76 & 0.87 & 0.87 & 0.87 & 0.87 & 0.87 & 0.87 & \cellcolor[gray]{0.90}0.87 & \textbf{0.87} & 90 \\
\rule{0pt}{2.35ex}Population near hospital (\%) & 31.6 & \cellcolor[gray]{0.90}40.5 & \cellcolor[gray]{0.90}46.7 & \cellcolor[gray]{0.90}52.1 & 46.7 & 49.8 & \cellcolor[gray]{0.90}48.9 & \cellcolor[gray]{0.90}49.8 & \cellcolor[gray]{0.90}44.3 & \cellcolor[gray]{0.90}56.8 & \cellcolor[gray]{0.90}52.7 & \textbf{48.3} & 76 \\
\rule{0pt}{2.35ex}Everyday-service POIs (\%) & 65.4 & 80.0 & 80.0 & \cellcolor[gray]{0.90}66.8 & 80.0 & 80.0 & 80.0 & 80.0 & 80.0 & 80.0 & 80.0 & \textbf{80.0} & 90 \\
\midrule
\multicolumn{14}{@{}l}{\textbf{Economy, Function \& Centrality}} \\
\rule{0pt}{2.35ex}Non-residential volume (\%) & 3.0 & 23.8 & 22.6 & \cellcolor[gray]{0.90}18.5 & 23.5 & 23.7 & 23.4 & 23.5 & 22.9 & 22.9 & 23.3 & \textbf{23.3} & 89 \\
\rule{0pt}{2.35ex}Mean economic-output surface (millions) & 33.9 & \cellcolor[gray]{0.90}58.1 & \cellcolor[gray]{0.90}74.3 & \cellcolor[gray]{0.90}44.4 & \cellcolor[gray]{0.90}67.3 & \cellcolor[gray]{0.90}68.7 & \cellcolor[gray]{0.90}44.3 & \cellcolor[gray]{0.90}60.0 & \cellcolor[gray]{0.90}34.2 & \cellcolor[gray]{0.90}55.6 & \cellcolor[gray]{0.90}36.0 & \textbf{--} & -- \\
\rule{0pt}{2.35ex}Functional/centre area ratio & 2.89 & \cellcolor[gray]{0.90}5.75 & 5.24 & \cellcolor[gray]{0.90}4.42 & 6.52 & 5.87 & \cellcolor[gray]{0.90}5.45 & 6.36 & 5.92 & 6.58 & 5.90 & \textbf{6.06} & 86 \\
\rule{0pt}{2.35ex}Effective POI categories & 7.6 & \cellcolor[gray]{0.90}7.1 & \cellcolor[gray]{0.90}8.5 & \cellcolor[gray]{0.90}7.8 & \cellcolor[gray]{0.90}8.2 & \cellcolor[gray]{0.90}7.9 & \cellcolor[gray]{0.90}5.6 & \cellcolor[gray]{0.90}7.3 & \cellcolor[gray]{0.90}4.6 & \cellcolor[gray]{0.90}8.6 & \cellcolor[gray]{0.90}7.9 & \textbf{--} & -- \\
\midrule
\multicolumn{14}{@{}l}{\textbf{Culture, Institutions \& Identity}} \\
\rule{0pt}{2.35ex}National-capital share (\%) & 1.0 & 3.1 & 3.1 & \cellcolor[gray]{0.90}0.0 & \cellcolor[gray]{0.90}3.2 & \cellcolor[gray]{0.90}3.0 & 4.2 & 5.0 & 2.6 & 1.8 & \cellcolor[gray]{0.90}4.2 & \textbf{3.1} & -- \\
\rule{0pt}{2.35ex}Nearest UNESCO property (km) & 128 & \cellcolor[gray]{0.90}31 & \cellcolor[gray]{0.90}31 & \cellcolor[gray]{0.90}208 & \cellcolor[gray]{0.90}31 & \cellcolor[gray]{0.90}31 & \cellcolor[gray]{0.90}33 & \cellcolor[gray]{0.90}31 & \cellcolor[gray]{0.90}39 & \cellcolor[gray]{0.90}57 & \cellcolor[gray]{0.90}31 & \textbf{--} & -- \\
\rule{0pt}{2.35ex}Civic/cultural POIs (\%) & 1.3 & 3.7 & 3.7 & \cellcolor[gray]{0.90}0.0 & 3.7 & 3.7 & 3.7 & 3.7 & 2.9 & 3.7 & 3.7 & \textbf{3.7} & 90 \\
\bottomrule
\end{tabular*}
\endgroup

\caption{Expanded Table~1, part B: dimensions 21--40. Column definitions and
gray-cell semantics match Table~\ref{tab:parameter-a}. Dashes mark binary
weighted-share rows and dimensions without a reliable checkpoint.}
\label{tab:parameter-b}
\end{table}

\FloatBarrier

\begin{figure}[!htbp]
\centering
\includegraphics[width=.99\textwidth]{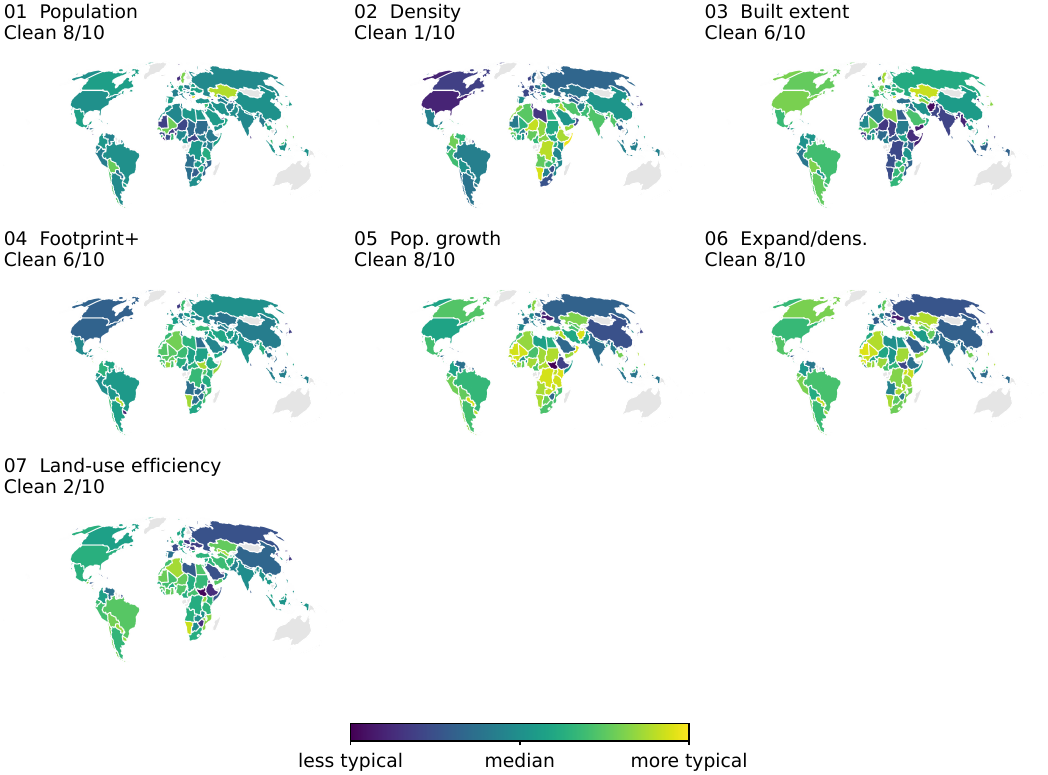}
\caption{Expanded Figure~2A: all seven \emph{Scale, Extent \& Change}
dimensions.
``Clean $k$/10'' gives the number of checkpoints used. Gray countries indicate
that the row-specific frame contains no observed city. Country fills summarize observed
urban-centre ranks, medianed within dependency components and then across
components.}
\label{fig:maps-a}
\end{figure}

\begin{figure}[!htbp]
\centering
\includegraphics[width=.99\textwidth]{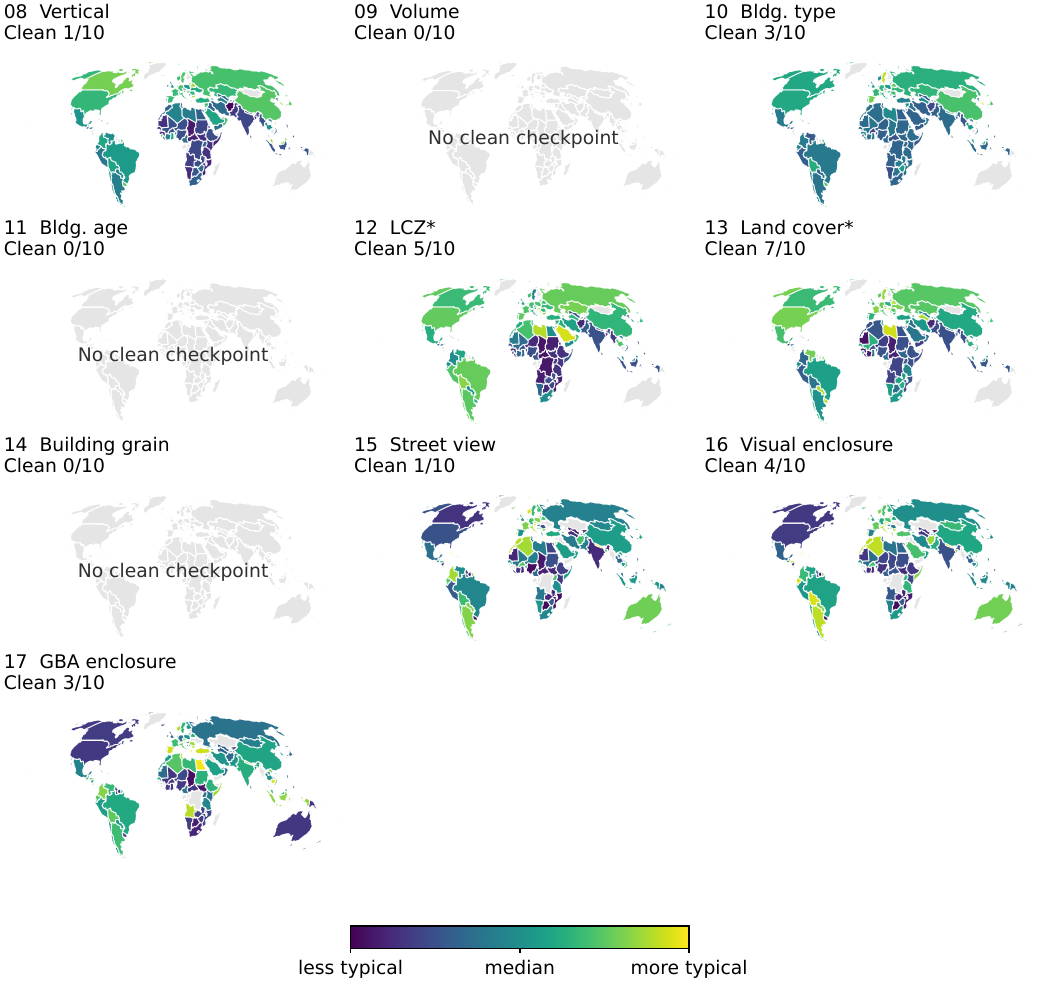}
\caption{Expanded Figure~2B: all ten \emph{Built Form \& Land Use}
dimensions. Blank
panels mark dimensions without a reliable checkpoint.}
\label{fig:maps-b}
\end{figure}

\begin{figure}[!htbp]
\centering
\includegraphics[width=.99\textwidth]{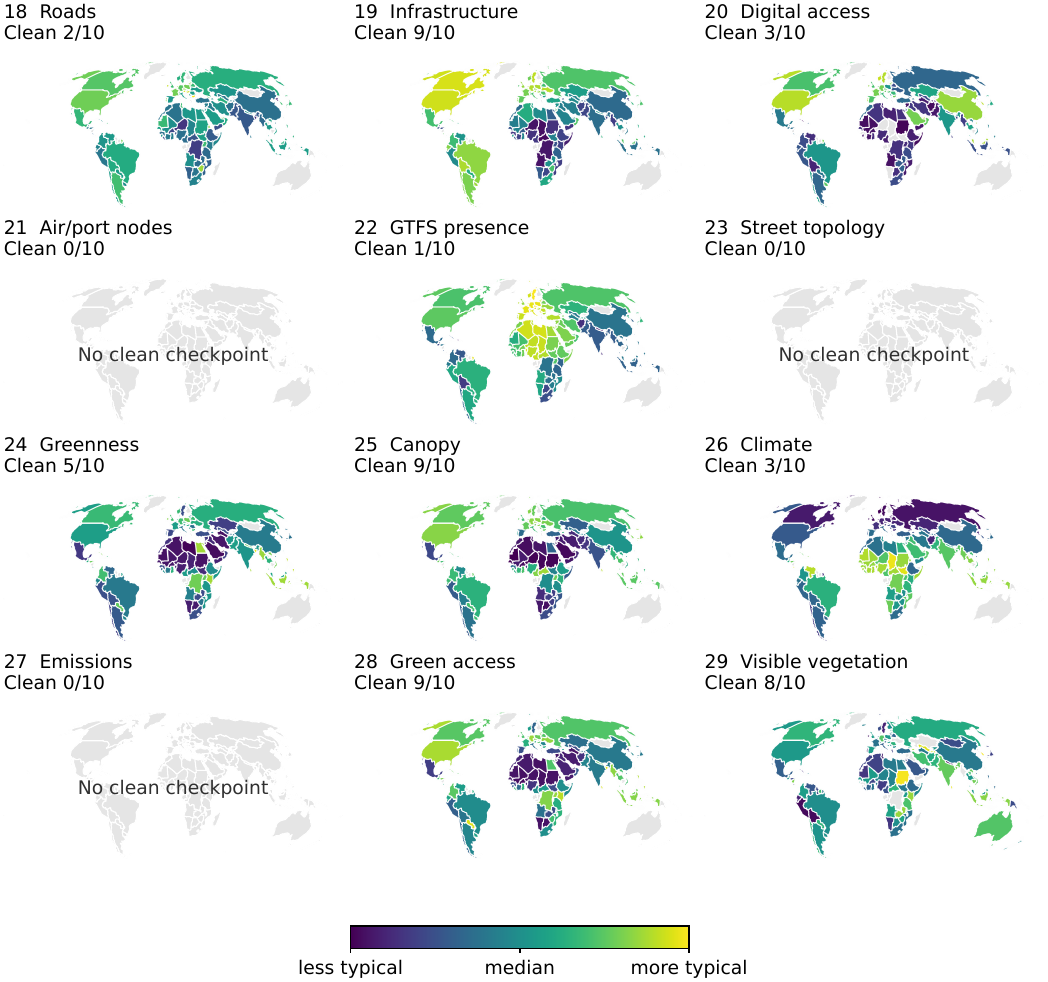}
\caption{Expanded Figure~2C: all six \emph{Networks, Mobility \&
Infrastructure} and six \emph{Ecology \& Environment} dimensions.
Percentiles are normalized within each model and row before
dependency-balanced aggregation.}
\label{fig:maps-c}
\end{figure}

\begin{figure}[!htbp]
\centering
\includegraphics[width=.99\textwidth]{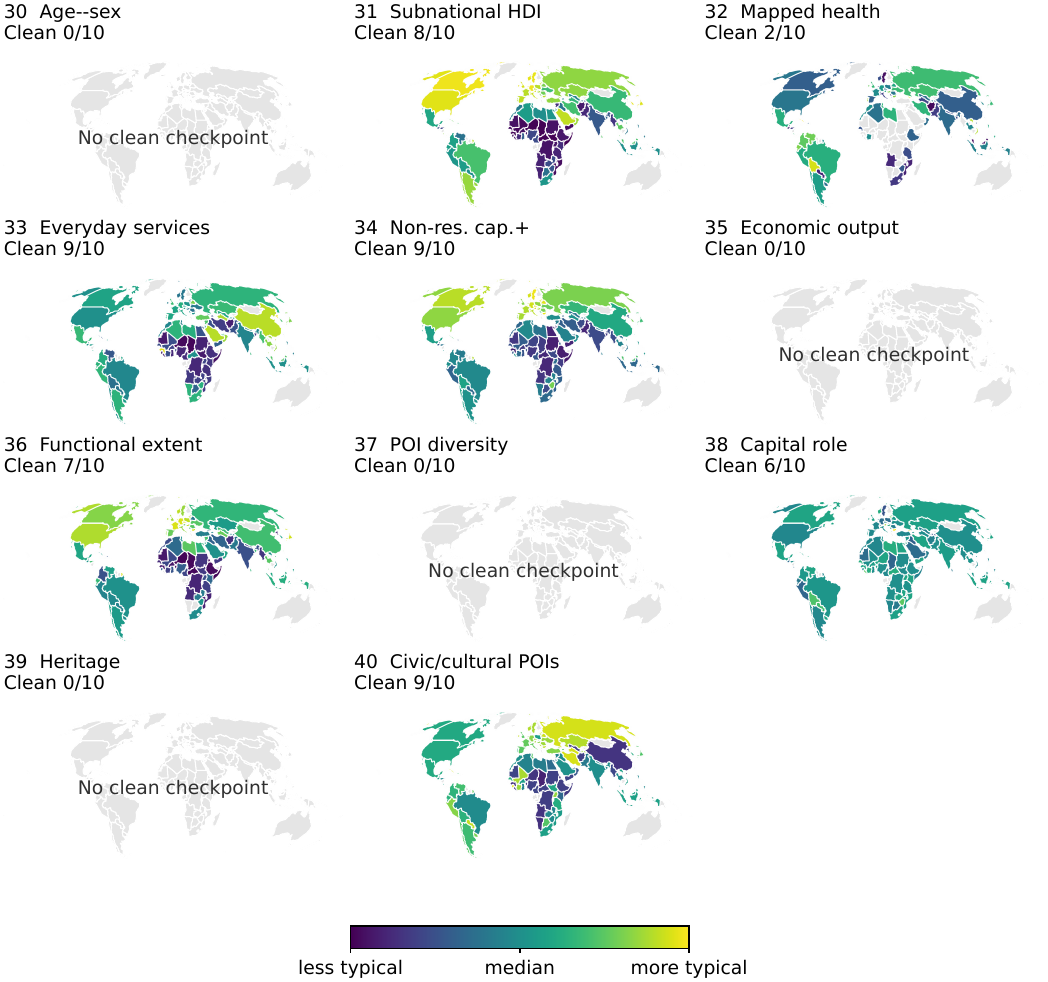}
\caption{Expanded Figure~2D: all \emph{Population, Housing \& Public Life},
\emph{Economy, Function \& Centrality}, and \emph{Culture, Institutions \&
Identity} dimensions. The figure retains availability-frame gaps, with
spatial interpolation omitted.}
\label{fig:maps-d}
\end{figure}

\FloatBarrier

\begin{figure}[!htbp]
\centering
\includegraphics[height=.82\textheight,keepaspectratio]{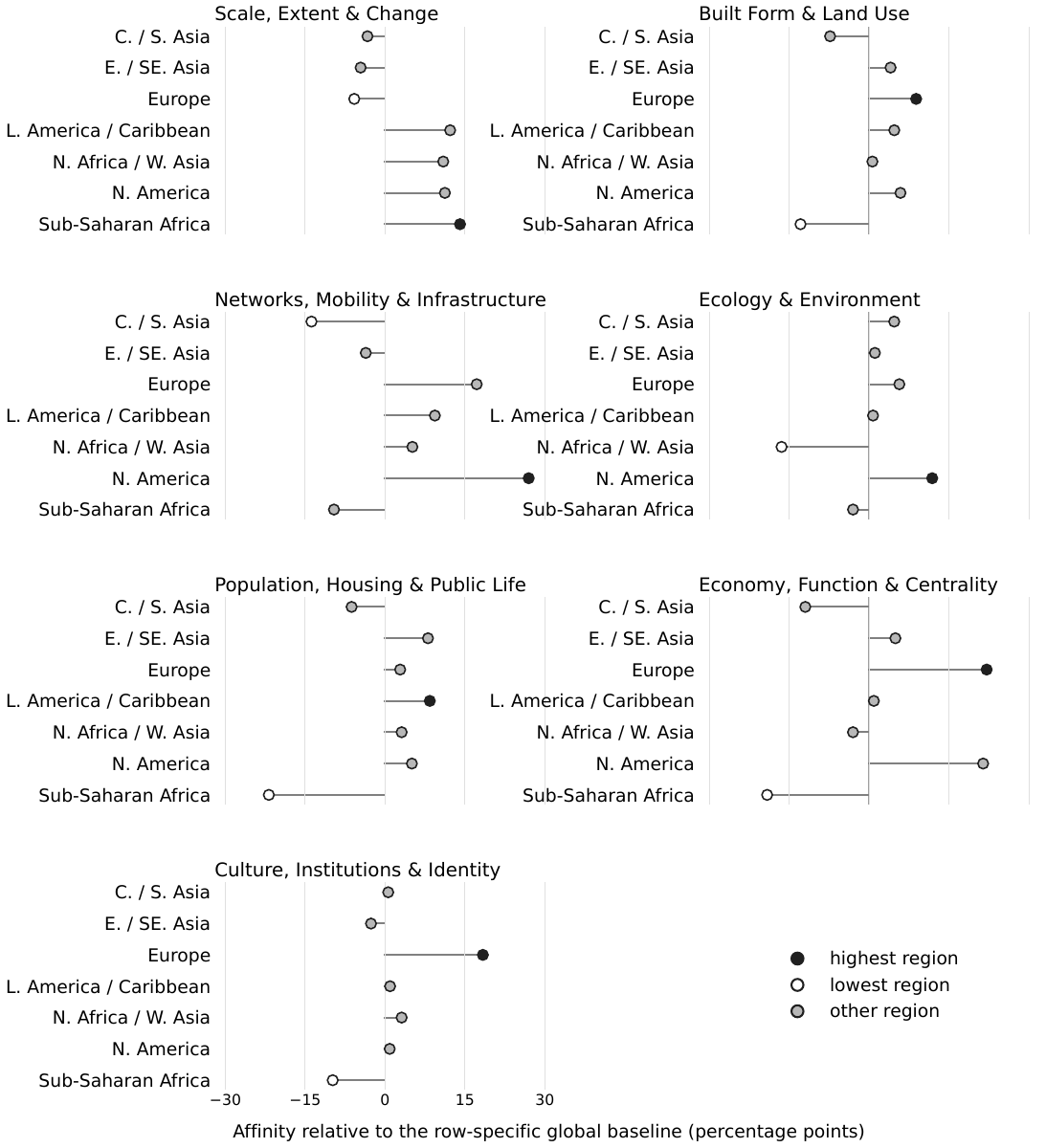}
\caption{Complete seven-domain regional affinity. Each point is the
dependency-balanced affinity relative to the row-specific global baseline,
in percentage points. Filled and open points mark the highest and lowest
region within a domain; gray points retain all other regions. The overall
Europe/Northern-America tilt contrasts with \emph{Scale, Extent \& Change},
where Sub-Saharan Africa and Latin America and the Caribbean rank highest.
Table~\ref{tab:regional-grids} reports the exact values.}
\label{fig:regional-expanded}
\end{figure}

\begin{table}[!htbp]
\centering
\begingroup
\normalsize
\setlength{\tabcolsep}{3.2pt}
\renewcommand{\arraystretch}{1.75}
\begin{tabular*}{\textwidth}{@{\extracolsep{\fill}}lrrrrrrr@{}}
\toprule
Summary & \shortstack{C./S.\\Asia} & \shortstack{E./SE.\\Asia} & Europe & LAC & \shortstack{N. Afr./\\W. Asia} & \shortstack{N.\\America} & SSA \\
\midrule
Overall & -6.5 & +1.9 & +10.8 & +5.0 & +3.0 & +10.0 & -10.2 \\
\midrule
Scale, Extent \& Change & -3.3 & -4.6 & -5.8 & +12.2 & +10.9 & +11.2 & +14.1 \\
Built Form \& Land Use & -7.3 & +4.0 & +8.8 & +4.7 & +0.6 & +5.9 & -12.9 \\
Networks, Mobility \& Infrastructure & -13.8 & -3.6 & +17.2 & +9.3 & +5.1 & +27.0 & -9.6 \\
Ecology \& Environment & +4.7 & +1.1 & +5.7 & +0.7 & -16.4 & +11.9 & -3.0 \\
Population, Housing \& Public Life & -6.3 & +8.1 & +2.8 & +8.4 & +3.1 & +5.0 & -21.8 \\
Economy, Function \& Centrality & -12.0 & +4.9 & +22.1 & +0.9 & -3.0 & +21.4 & -19.1 \\
Culture, Institutions \& Identity & +0.6 & -2.6 & +18.4 & +0.9 & +3.1 & +0.9 & -9.8 \\
\bottomrule
\end{tabular*}
\endgroup

\par\medskip
\centering
\begingroup
\normalsize
\setlength{\tabcolsep}{3.2pt}
\renewcommand{\arraystretch}{1.75}
\begin{tabular*}{\textwidth}{@{\extracolsep{\fill}}lrrrrrrr@{}}
\toprule
Checkpoint & \shortstack{C./S.\\Asia} & \shortstack{E./SE.\\Asia} & Europe & LAC & \shortstack{N. Afr./\\W. Asia} & \shortstack{N.\\America} & SSA \\
\midrule
Qwen3 8B & -6.9 & +1.7 & +13.4 & +5.7 & +3.0 & +10.7 & -10.2 \\
Qwen2.5 7B & -8.5 & +2.4 & +10.8 & +5.4 & +3.1 & +12.5 & -8.8 \\
DeepSeek-R1 7B & +0.3 & +2.2 & +2.5 & +3.9 & +0.0 & -1.6 & -5.8 \\
Gemma2 9B & -6.5 & +3.1 & +11.4 & +4.1 & +3.1 & +9.2 & -10.3 \\
Gemma3 12B & -7.4 & +1.2 & +9.4 & +4.1 & +2.9 & +10.9 & -10.2 \\
Llama3.1 8B & -5.9 & +1.9 & +8.8 & +5.6 & +3.7 & +7.8 & -10.2 \\
Mistral Nemo 12B & -7.0 & +1.7 & +10.0 & +5.0 & +3.1 & +5.3 & -11.4 \\
OLMo2 7B & -6.5 & +1.6 & +12.7 & +4.8 & +1.4 & +10.8 & -9.5 \\
Phi-4 14B & -4.2 & +1.4 & +10.8 & +5.2 & +1.0 & +8.8 & -10.8 \\
GLM-4 9B & -6.1 & +2.2 & +11.1 & +2.5 & +2.6 & +14.3 & -10.2 \\
\bottomrule
\end{tabular*}
\endgroup

\caption{Complete regional affinity grids. The upper panel gives
dependency-balanced domain-by-region values; Overall gives the seven domains
equal weight. The lower panel gives checkpoint-by-region values after
medianing dimensions within domains and weighting the seven domains equally.
LAC is Latin America and the Caribbean; SSA is Sub-Saharan Africa.}
\label{tab:regional-grids}
\end{table}

\FloatBarrier

\begin{table}[!htbp]
\centering
\begingroup
\footnotesize
\setlength{\tabcolsep}{2.05pt}
\renewcommand{\arraystretch}{1.80}
\begin{tabular*}{\textwidth}{@{\extracolsep{\fill}}>{\raggedright\arraybackslash}p{.265\textwidth}rrrrrrrrrr@{}}
\toprule
Dimension & Q3 & Q2.5 & DS & G2 & G3 & L3.1 & MN & O2 & P4 & GLM \\
\midrule
\multicolumn{11}{@{}l}{\textbf{Scale, Extent \& Change}} \\
\rule{0pt}{2.35ex}Population & 6.9 & 8.2 & 45.2 & 6.8 & 10.6 & 8.7 & 7.3 & 12.3 & 7.7 & 8.0 \\
\rule{0pt}{2.35ex}Density & 24.4 & 14.6 & 19.8 & 19.7 & 13.7 & 27.1 & 13.8 & 20.6 & 14.7 & 17.1 \\
\rule{0pt}{2.35ex}Built extent & 6.6 & 7.7 & 20.6 & 3.8 & 4.8 & 7.3 & 3.9 & 15.5 & 4.2 & 7.7 \\
\rule{0pt}{2.35ex}Footprint+ & 15.0 & 10.5 & 22.1 & 16.7 & 12.6 & 9.3 & 19.6 & 16.3 & 11.9 & 21.9 \\
\rule{0pt}{2.35ex}Pop. growth & 9.1 & 4.0 & 29.0 & 3.2 & 3.9 & 4.5 & 2.8 & 7.1 & 3.5 & 44.4 \\
\rule{0pt}{2.35ex}Expand/dens. & 9.3 & 10.1 & 16.4 & 10.8 & 13.7 & 11.9 & 7.3 & 10.3 & 9.9 & 9.1 \\
\rule{0pt}{2.35ex}Land-use efficiency & 12.7 & 10.5 & 22.4 & 9.8 & 14.1 & 14.6 & 9.9 & 12.1 & 8.5 & 16.6 \\
\midrule
\multicolumn{11}{@{}l}{\textbf{Built Form \& Land Use}} \\
\rule{0pt}{2.35ex}Vertical & 9.8 & 12.1 & 29.0 & 9.0 & 8.6 & 15.2 & 11.7 & 31.7 & 8.5 & 16.0 \\
\rule{0pt}{2.35ex}Volume & 21.6 & 21.6 & 30.1 & 22.8 & 20.0 & 25.1 & 28.1 & 22.5 & 23.1 & 25.7 \\
\rule{0pt}{2.35ex}Bldg. type & 20.3 & 28.5 & 23.5 & 19.9 & 27.2 & 36.1 & 28.3 & 23.6 & 20.1 & 29.8 \\
\rule{0pt}{2.35ex}Bldg. age & 18.7 & 19.0 & 21.2 & 14.5 & 31.3 & 20.0 & 18.3 & 19.4 & 29.5 & 23.2 \\
\rule{0pt}{2.35ex}LCZ* & 11.9 & 14.4 & 21.3 & 16.4 & 23.2 & 12.3 & 11.3 & 15.8 & 16.8 & 18.0 \\
\rule{0pt}{2.35ex}Land cover* & 10.1 & 11.9 & 22.4 & 12.1 & 7.0 & 9.2 & 11.0 & 19.3 & 7.4 & 9.8 \\
\rule{0pt}{2.35ex}Building grain & 13.7 & 11.7 & 28.5 & 12.7 & 21.6 & 15.8 & 18.4 & 16.5 & 20.1 & 14.0 \\
\rule{0pt}{2.35ex}Street view & 17.1 & 16.6 & 24.5 & 10.9 & 11.3 & 16.5 & 15.1 & 15.7 & 14.1 & 21.4 \\
\rule{0pt}{2.35ex}Visual enclosure & 6.1 & 5.4 & 21.7 & 5.3 & 13.5 & 4.7 & 3.7 & 6.5 & 4.6 & 5.8 \\
\rule{0pt}{2.35ex}GBA enclosure & 9.1 & 10.8 & 20.1 & 14.9 & 43.7 & 11.1 & 11.7 & 22.2 & 9.7 & 16.2 \\
\midrule
\multicolumn{11}{@{}l}{\textbf{Networks, Mobility \& Infrastructure}} \\
\rule{0pt}{2.35ex}Roads & 13.1 & 12.8 & 24.9 & 11.2 & 22.3 & 15.5 & 22.3 & 19.9 & 12.0 & 17.6 \\
\rule{0pt}{2.35ex}Infrastructure & 9.6 & 8.5 & 11.3 & 5.9 & 7.8 & 7.0 & 5.6 & 9.9 & 6.0 & 8.0 \\
\rule{0pt}{2.35ex}Digital access & 5.9 & 7.8 & 23.7 & 5.6 & 5.1 & 8.2 & 4.7 & 8.9 & 5.8 & 20.5 \\
\bottomrule
\end{tabular*}
\endgroup

\caption{Expanded Table~2, part A: dimensions 1--20. The
leave-component-out comparison removes the focal checkpoint's entire
dependency component before constructing consensus. Values are mean absolute
profile-rank differences in percentage points. Parts A and B report all 400
dimension--checkpoint distances.}
\label{tab:divergence-a}
\end{table}

\begin{table}[!htbp]
\centering
\begingroup
\footnotesize
\setlength{\tabcolsep}{2.05pt}
\renewcommand{\arraystretch}{1.80}
\begin{tabular*}{\textwidth}{@{\extracolsep{\fill}}>{\raggedright\arraybackslash}p{.265\textwidth}rrrrrrrrrr@{}}
\toprule
Dimension & Q3 & Q2.5 & DS & G2 & G3 & L3.1 & MN & O2 & P4 & GLM \\
\midrule
\multicolumn{11}{@{}l}{\textbf{Networks, Mobility \& Infrastructure}} \\
\rule{0pt}{2.35ex}Air/port nodes & 22.3 & 16.8 & 31.8 & 12.9 & 13.0 & 18.9 & 10.6 & 15.6 & 17.8 & 17.8 \\
\rule{0pt}{2.35ex}GTFS presence & 11.7 & 12.0 & 13.4 & 11.5 & 13.7 & 12.8 & 14.8 & 11.7 & 16.6 & 10.7 \\
\rule{0pt}{2.35ex}Street topology & 16.3 & 22.3 & 26.1 & 13.2 & 16.5 & 12.4 & 13.1 & 14.6 & 15.3 & 21.5 \\
\midrule
\multicolumn{11}{@{}l}{\textbf{Ecology \& Environment}} \\
\rule{0pt}{2.35ex}Greenness & 13.8 & 8.5 & 17.9 & 7.9 & 17.3 & 18.0 & 15.7 & 15.3 & 9.8 & 11.9 \\
\rule{0pt}{2.35ex}Canopy & 11.1 & 8.7 & 24.0 & 12.5 & 7.9 & 9.1 & 8.1 & 17.9 & 8.4 & 12.2 \\
\rule{0pt}{2.35ex}Climate & 15.3 & 15.9 & 25.4 & 38.6 & 14.3 & 22.1 & 18.4 & 19.0 & 21.7 & 22.9 \\
\rule{0pt}{2.35ex}Emissions & 19.1 & 14.1 & 23.7 & 26.4 & 30.0 & 15.9 & 12.1 & 14.9 & 25.1 & 17.7 \\
\rule{0pt}{2.35ex}Green access & 2.6 & 3.2 & 20.0 & 2.0 & 3.3 & 4.7 & 1.7 & 2.9 & 2.6 & 3.0 \\
\rule{0pt}{2.35ex}Visible vegetation & 8.3 & 9.3 & 33.5 & 9.3 & 26.1 & 8.5 & 8.3 & 8.0 & 11.0 & 14.3 \\
\midrule
\multicolumn{11}{@{}l}{\textbf{Population, Housing \& Public Life}} \\
\rule{0pt}{2.35ex}Age--sex & 15.0 & 36.9 & 33.6 & 18.0 & 17.1 & 24.4 & 20.4 & 18.2 & 27.2 & 24.5 \\
\rule{0pt}{2.35ex}Subnational HDI & 3.3 & 2.8 & 22.7 & 1.8 & 3.6 & 4.1 & 2.6 & 6.4 & 2.7 & 5.0 \\
\rule{0pt}{2.35ex}Mapped health & 12.1 & 9.5 & 17.0 & 9.1 & 12.1 & 14.7 & 8.7 & 16.7 & 13.1 & 12.6 \\
\rule{0pt}{2.35ex}Everyday services & 5.4 & 3.8 & 27.0 & 3.8 & 4.6 & 7.5 & 3.8 & 6.7 & 3.2 & 7.5 \\
\midrule
\multicolumn{11}{@{}l}{\textbf{Economy, Function \& Centrality}} \\
\rule{0pt}{2.35ex}Non-res. cap.+ & 4.0 & 6.4 & 15.6 & 3.0 & 3.9 & 5.3 & 2.3 & 2.7 & 3.4 & 2.9 \\
\rule{0pt}{2.35ex}Economic output & 17.3 & 23.0 & 32.6 & 17.6 & 19.1 & 21.6 & 21.3 & 25.5 & 20.9 & 27.0 \\
\rule{0pt}{2.35ex}Functional extent & 9.8 & 13.0 & 19.3 & 4.9 & 11.4 & 10.1 & 5.5 & 10.7 & 8.7 & 8.9 \\
\rule{0pt}{2.35ex}POI diversity & 25.1 & 20.7 & 24.0 & 18.4 & 16.2 & 22.5 & 16.8 & 28.2 & 20.0 & 17.2 \\
\midrule
\multicolumn{11}{@{}l}{\textbf{Culture, Institutions \& Identity}} \\
\rule{0pt}{2.35ex}Capital role & 26.0 & 25.9 & 27.7 & 26.6 & 26.1 & 26.5 & 26.3 & 26.3 & 25.8 & 26.4 \\
\rule{0pt}{2.35ex}Heritage & 11.3 & 7.6 & 33.9 & 9.3 & 11.7 & 13.8 & 9.1 & 11.6 & 21.4 & 12.0 \\
\rule{0pt}{2.35ex}Civic/cultural POIs & 6.1 & 7.4 & 42.9 & 5.4 & 7.3 & 8.6 & 5.8 & 10.3 & 6.2 & 7.0 \\
\bottomrule
\end{tabular*}
\endgroup

\caption{Expanded Table~2, part B: dimensions 21--40.}
\label{tab:divergence-b}
\end{table}

\FloatBarrier

\begin{figure}[!htbp]
\centering
\includegraphics[width=.99\textwidth]{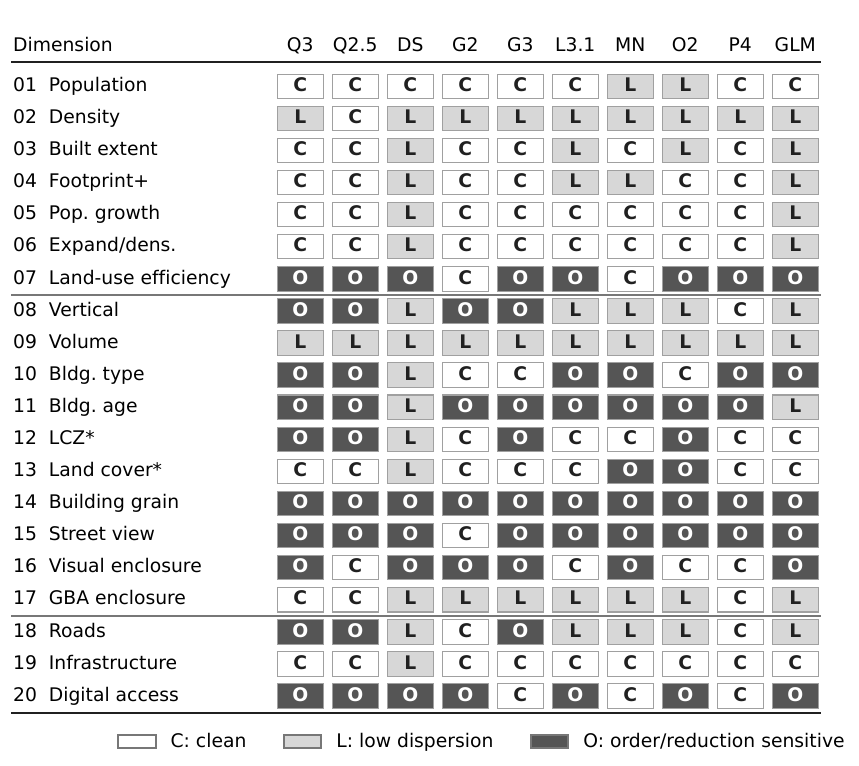}
\caption{Complete measurement-status matrix, dimensions 1--20. C denotes
clean measurement, L low score dispersion, and O ordering or reduced-form
sensitivity. Horizontal rules mark the seven-domain boundaries. All labels
describe measurement reliability. The complete matrix contains all 400
typicality cells: 162 are clean, 72 have
low dispersion, and 166 are ordering or reduction sensitive. Gray parameter
cells and blank identified maps in earlier displays follow directly from this
matrix.}
\label{fig:status-a}
\end{figure}

\begin{figure}[!htbp]
\centering
\includegraphics[width=.99\textwidth]{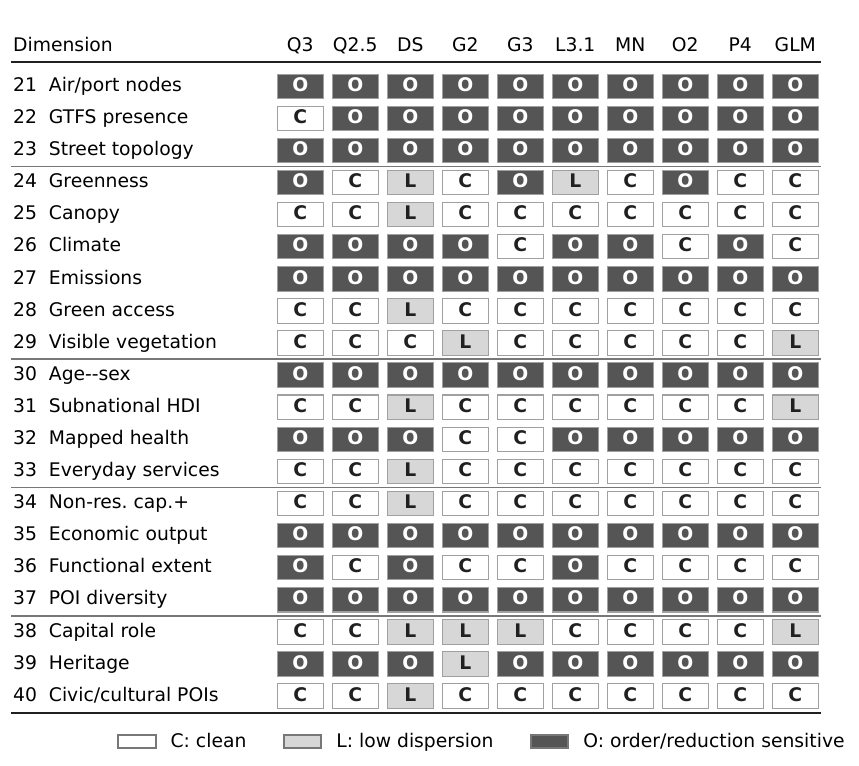}
\caption{Complete measurement-status matrix, dimensions 21--40. Model
abbreviations follow the main paper and Tables~\ref{tab:parameter-a}--
\ref{tab:parameter-b}; horizontal rules retain the same seven-domain
organization. The replication sample retains 285/291 city-count directions
and has effect-rank correlation .982.}
\label{fig:status-b}
\end{figure}

\FloatBarrier

\section{Complete Joint-Profile Results}

The complete-case frame contains all 48 centres with a value for the
designated parameter of every dimension. It spans 29 countries and six world
regions; Central and Southern Asia is absent. Each anonymous card contains
exactly 40 physical-unit lines and is rated in forward and reverse order by
all ten checkpoints. The dimension-wise predictor first midranks profiles
within checkpoint and dimension, then means within the seven domains and
gives the domains equal weight.

\begin{table}[h!]
\centering
\small
\setlength{\tabcolsep}{4pt}
\begin{tabular}{@{}lrrr@{}}
\toprule
Checkpoint & Marginal--joint $\rho$ & 95\% CI & Order $\rho$ \\
\midrule
G2 & 0.700 & [0.469, 0.869] & 0.906 \\
Q2.5 & 0.688 & [0.461, 0.835] & 0.502 \\
G3 & 0.684 & [0.464, 0.823] & 0.832 \\
O2 & 0.666 & [0.423, 0.834] & 0.746 \\
P4 & 0.660 & [0.423, 0.823] & 0.919 \\
Q3 & 0.500 & [0.230, 0.708] & 0.895 \\
DS & 0.500 & [0.291, 0.660] & 0.629 \\
MN & 0.468 & [0.205, 0.674] & 0.875 \\
L3.1 & 0.376 & [0.062, 0.628] & 0.860 \\
GLM & 0.258 & [-0.013, 0.501] & 0.892 \\
\midrule
Overall & 0.500 & [0.378, 0.695] & -- \\
\bottomrule
\end{tabular}

\caption{Complete checkpoint results underlying Figure~4. Marginal--joint
$\rho$ compares the seven-domain-balanced dimension-wise predictor with the
direct 40-indicator score. Order $\rho$ compares forward and reverse cards.
Checkpoint intervals and the dependency-balanced Overall interval use 5,000
city-bootstrap replicates.}
\label{tab:joint-profile}
\end{table}

All 960 ratings are valid. The dependency-balanced statistic is .500
(95\% CI [.378, .695]); all seven components are positive. Omitting any one
domain yields .350--.675. These estimates apply to the 48-city complete-case
frame.

\FloatBarrier

\section{Sensitivity Analyses}

\ifdefined\ArxivIntegratedAppendix\else
\begin{table}[!t]
\centering
\small
\begin{tabular}{@{}lrrr@{}}
\toprule
Analysis & $n$ & Rank $\rho$ & Shift (pp) \\
\midrule
Upper tail: 30\% & 155 & 0.899 & 4.7 \\
Upper tail: 10\% & 155 & 0.908 & 5.0 \\
Checkpoint median & 28 & 0.990 & 0.0 \\
Component mean & 28 & 0.943 & 0.1 \\
Leave one component out & 12 & 0.965 & 0.5 \\
Chinese instructions & 48 & 0.950 & 4.2 \\
\bottomrule
\end{tabular}

\caption{Sensitivity of the standardized 40-dimensional portrait. Rank
$\rho$ compares each alternative with the primary analysis. Shift is the
median absolute change in source-percentile rank; the leave-one-component row
reports the least favorable correlation and largest shift across the seven
omissions.}
\label{tab:sensitivity}
\end{table}
\fi

Changing the upper typicality tail from 20\% to 30\% or 10\% yields rank
correlations of .899 and .908 across 155 comparable clean
model--dimension cells. Checkpoint-median, component-mean, and
leave-one-component analyses yield cross-dimensional correlations of
.990, .943, and at least .965. All aggregation variants keep every comparable
dimension on the same side of the global median. Replacing the joint-profile
instructions with Chinese yields a component-balanced correlation of .950
(95\% city-bootstrap CI [.907, .962]); the median checkpoint correlation is
.945.

\begin{table}[!t]
\centering
\small
\begingroup
\setlength{\tabcolsep}{2.5pt}
\renewcommand{\arraystretch}{1.12}
\begin{tabular}{@{}lrrrrr@{}}
\toprule
Region & Raw & Pop. & Scale & HDI & S+H \\
\midrule
C. / S. Asia & -11.4 & -10.7 & -4.1 & -0.8 & -0.6 \\
E. / SE. Asia & +3.8 & +3.4 & +3.4 & +1.7 & +2.0 \\
Europe & +14.1 & +15.2 & +10.2 & +4.0 & +4.2 \\
L. America / Carib. & +3.8 & +3.8 & +2.4 & +2.0 & +2.0 \\
N. Africa / W. Asia & +0.6 & +0.3 & +3.0 & +0.9 & +0.0 \\
N. America & +10.3 & +10.1 & +1.5 & -0.6 & -2.3 \\
Sub-Saharan Africa & -17.5 & -18.3 & -12.0 & -1.5 & -3.5 \\
\bottomrule
\end{tabular}
\endgroup

\caption{Scale and development conditioning underlying Figure~3B. Values are
dependency-balanced regional affinities in percentage points relative to the
row-specific global baseline across 123 reliable non-scale model--dimension
cells. Pop. adjusts population; Scale jointly adjusts population, built
surface, and density; HDI adjusts subnational HDI; Scale+HDI includes all four
covariates.}
\label{tab:regional-controls}
\end{table}

Population adjustment leaves the region ordering unchanged
($\rho=1.000$ with the unadjusted synthesis; median absolute shift $0.4$
points). Scale adjustment changes the ordering to $\rho=.679$ and shifts
regions by a median $3.9$ points; HDI and Scale+HDI yield $\rho=.786$ and
$.607$, with median shifts of $10.1$ and $9.9$ points. The median
cell-level weighted $R^2$ is .02 for population, .17 for scale, .21 for HDI,
and .28 for Scale+HDI. Minimum retained coverage is 98.2\%.

\FloatBarrier
\section{Prompts and Reproducibility}

This section records checkpoint pins, prompt wording, inference settings, and
artifact receipts for the 40-dimensional matrix and validation analyses.

\begin{table}[!htbp]
\centering
\begingroup
\small
\setlength{\tabcolsep}{3pt}
\renewcommand{\arraystretch}{1.75}
\begin{tabular*}{\textwidth}{@{\extracolsep{\fill}}p{.13\textwidth}p{.29\textwidth}p{.10\textwidth}p{.19\textwidth}cc@{}}
\toprule
Checkpoint & Repository & Revision & Dependency component & TP & Mode \\
\midrule
Qwen3 8B & Qwen/Qwen3-8B & \texttt{b968826d} & qwen\_\allowbreak{}deepseek\_\allowbreak{}lineage & 1 & non-thinking \\
Qwen2.5 7B & Qwen/Qwen2.5-7B-Instruct & \texttt{a09a3545} & qwen\_\allowbreak{}deepseek\_\allowbreak{}lineage & 1 & chat \\
DeepSeek-R1 7B & deepseek-ai/DeepSeek-R1-Distill-Qwen-7B & \texttt{916b56a4} & qwen\_\allowbreak{}deepseek\_\allowbreak{}lineage & 1 & chat \\
Gemma2 9B & google/gemma-2-9b-it & \texttt{11c9b309} & gemma & 1 & chat \\
Gemma3 12B & google/gemma-3-12b-it & \texttt{96b6f1ec} & gemma & 2 & chat \\
Llama3.1 8B & meta-llama/Llama-3.1-8B-Instruct & \texttt{0e9e39f2} & llama31 & 1 & chat \\
Mistral Nemo 12B & mistralai/Mistral-Nemo-Instruct-2407 & \texttt{04d8a905} & mistral\_nemo & 2 & chat \\
OLMo2 7B & allenai/OLMo-2-1124-7B-Instruct & \texttt{470b1fba} & olmo2 & 1 & chat \\
Phi-4 14B & microsoft/phi-4 & \texttt{2db69c1c} & phi4 & 2 & chat \\
GLM-4 9B & zai-org/GLM-4-9B-0414 & \texttt{645b8482} & glm4 & 1 & chat \\
\bottomrule
\end{tabular*}
\endgroup

\par\medskip
\begin{minipage}[t]{.48\textwidth}
\textbf{Typicality prompt.}
Each anonymous profile instantiates the same template. Multi-field rows use a
hash-selected cyclic order and its exact reverse; geographic and source
identifiers are absent.\par\smallskip

\textbf{Anonymous measured city dimension (\emph{dimension label}):}\\
\emph{field-value lines in physical units}\\[2pt]
Judge conceptual typicality only. Do not judge quality, wealth, safety,
sustainability, functioning, or desirability. Based only on this measured
dimension, how typical would a city with this condition be as an example of
what people generally mean by a city? Use a 1-to-7 scale where 1 means not at
all typical and 7 means extremely typical. Return exactly one digit from 1 to
7 and nothing else.
\end{minipage}
\hfill
\begin{minipage}[t]{.48\textwidth}
\textbf{Matched diagnostic and completion.}
The matched diagnostic replaces the instruction with: ``Judge general
desirability as a place to live only. Do not judge whether the condition is
typical or representative of cities.'' It uses the same anonymous fields,
order controls, scale, and model pins. The joint validation retains the
typicality instruction but replaces the dimension heading with ``Anonymous
measured city profile (40 urban indicators)'' and asks for the overall profile
to be considered jointly.\par\smallskip

Across the complete matrix, joint-profile validation, and language
sensitivity, each checkpoint completed 127,298 ratings, for 1,272,980 valid
records overall, with zero external model API calls. The appendix generator
additionally binds the
40-row parameter ledger, 400 raw-unit model values, clean-only map inputs,
four global map pages, seven-domain forest figure, joint-profile forest and
table, two divergence tables, and two status-matrix pages.
\end{minipage}
\caption{Model pins and prompt record. Revision is the first eight
characters of the repository revision; TP is tensor parallelism.
Qwen-derived and Gemma checkpoints share dependency components, leaving seven
components.}
\label{tab:model-roster}
\end{table}

\ifdefined\ArxivIntegratedAppendix
\begin{table}[!htbp]
\centering
\begingroup
\renewcommand{\arraystretch}{1.50}
\begin{tabular}{@{}p{.22\textwidth}p{.72\textwidth}@{}}
\toprule
Setting & Value \\
\midrule
Rating scale & Structured choice 1--7 \\
Generation & temperature 0; seed 0; max tokens 1 \\
Score & Expected rating from normalized choice log-probabilities \\
Constructs & Typicality and matched desirability \\
Field-order control & Forward/reverse for multi-field profiles \\
40D matrix & 127,106 ratings per checkpoint \\
Joint/language checks & 192 ratings per checkpoint \\
Completed ratings & 1,272,980 \\
Runtime & vLLM; bfloat16; max context 1024 \\
Hardware & two hosts; 4 x NVIDIA GeForce RTX 4090 each \\
Retries/timeout & 3 / 120 s \\
External APIs & 0 calls; offline, hash-bound weights \\
Dependency aggregation & Median within component, then across 7 components \\
Survey inference & Design weights; 1,000 bootstrap replicates; seeds 20260724/20260725 \\
\bottomrule
\end{tabular}
\endgroup

\caption{Inference and aggregation settings for the complete audit.}
\label{tab:runtime-settings}
\end{table}

\begin{table}[!htbp]
\centering
\begingroup
\renewcommand{\arraystretch}{1.50}
\begin{tabular}{@{}p{.52\textwidth}p{.12\textwidth}p{.28\textwidth}@{}}
\toprule
Artifact & Status & SHA-256 prefix \\
\midrule
40 x 10 matrix & pass & \texttt{70b4e2cb99c9} \\
Raw-parameter table/maps & pass & \texttt{1eb02bc9473a} \\
Regional affinity & pass & \texttt{18b0662f9043} \\
Scale-adjusted regional affinity & pass & \texttt{b35d84c77f83} \\
Model divergence & pass & \texttt{90aef8f273fa} \\
Full Overture audit & pass & \texttt{6987103b68ba} \\
Joint-profile analysis & pass & \texttt{5a465b819d3a} \\
Sensitivity analysis & pass & \texttt{77a45bb38698} \\
Chinese-instruction analysis & pass & \texttt{f3129f657f4b} \\
\bottomrule
\end{tabular}
\endgroup

\par\medskip
\begin{minipage}{.96\textwidth}
\small
\textbf{Checklist scope.}
The paper is empirical and introduces neither a theoretical contribution nor
a new upstream dataset. The accompanying Code and Data Supplement contains
preprocessing, inference, analysis, and display code, frozen configurations,
non-restricted derived artifacts, and verification instructions. Third-party
source payloads, model weights, and restricted raw responses are excluded;
public identifiers and hashes are retained. Development-range and
infrastructure items are partial because settings were fixed rather than
tuned and CPU, host-memory, and operating-system details were not recorded.
The statistical-testing item is partial because inference uses design-aware
bootstrap intervals and replication rather than a performance-improvement
hypothesis test. Source code will be released publicly upon publication.
\end{minipage}
\caption{Receipt-bound artifacts. Prefixes identify archived files; full
hashes and all bound input/output hashes remain in the repository artifacts.}
\label{tab:runtime-receipts}
\end{table}
\else
\begin{table}[!htbp]
\centering
\textbf{Inference and aggregation.}\par\smallskip
\begingroup
\renewcommand{\arraystretch}{1.50}
\begin{tabular}{@{}p{.22\textwidth}p{.72\textwidth}@{}}
\toprule
Setting & Value \\
\midrule
Rating scale & Structured choice 1--7 \\
Generation & temperature 0; seed 0; max tokens 1 \\
Score & Expected rating from normalized choice log-probabilities \\
Constructs & Typicality and matched desirability \\
Field-order control & Forward/reverse for multi-field profiles \\
40D matrix & 127,106 ratings per checkpoint \\
Joint/language checks & 192 ratings per checkpoint \\
Completed ratings & 1,272,980 \\
Runtime & vLLM; bfloat16; max context 1024 \\
Hardware & two hosts; 4 x NVIDIA GeForce RTX 4090 each \\
Retries/timeout & 3 / 120 s \\
External APIs & 0 calls; offline, hash-bound weights \\
Dependency aggregation & Median within component, then across 7 components \\
Survey inference & Design weights; 1,000 bootstrap replicates; seeds 20260724/20260725 \\
\bottomrule
\end{tabular}
\endgroup

\par\medskip
\textbf{Receipt-bound artifacts.}\par\smallskip
\begingroup
\renewcommand{\arraystretch}{1.50}
\begin{tabular}{@{}p{.52\textwidth}p{.12\textwidth}p{.28\textwidth}@{}}
\toprule
Artifact & Status & SHA-256 prefix \\
\midrule
40 x 10 matrix & pass & \texttt{70b4e2cb99c9} \\
Raw-parameter table/maps & pass & \texttt{1eb02bc9473a} \\
Regional affinity & pass & \texttt{18b0662f9043} \\
Scale-adjusted regional affinity & pass & \texttt{b35d84c77f83} \\
Model divergence & pass & \texttt{90aef8f273fa} \\
Full Overture audit & pass & \texttt{6987103b68ba} \\
Joint-profile analysis & pass & \texttt{5a465b819d3a} \\
Sensitivity analysis & pass & \texttt{77a45bb38698} \\
Chinese-instruction analysis & pass & \texttt{f3129f657f4b} \\
\bottomrule
\end{tabular}
\endgroup

\par\medskip
\begin{minipage}{.96\textwidth}
\small
\textbf{Checklist scope.}
The paper is empirical and introduces neither a theoretical contribution nor
a new upstream dataset. The accompanying Code and Data Supplement contains
preprocessing, inference, analysis, and display code, frozen configurations,
non-restricted derived artifacts, and verification instructions. Third-party
source payloads, model weights, and restricted raw responses are excluded;
public identifiers and hashes are retained. Development-range and
infrastructure items are partial because settings were fixed rather than
tuned and CPU, host-memory, and operating-system details were not recorded.
The statistical-testing item is partial because inference uses design-aware
bootstrap intervals and replication rather than a performance-improvement
hypothesis test. Source code will be released publicly upon publication.
\end{minipage}
\caption{Inference, aggregation, and receipt-bound artifacts. Receipt
prefixes identify archived files; full hashes and all bound input/output
hashes remain in the repository artifacts.}
\label{tab:runtime-receipts}
\end{table}
\fi
  \FloatBarrier
  \clearpage

\end{document}